\documentclass[10pt,twocolumn,letterpaper]{article}

\usepackage{cvpr}              

\usepackage{graphicx}
\usepackage{amsmath}
\usepackage{amssymb}
\usepackage{booktabs}
\usepackage{multirow}
\usepackage{makebox}
\usepackage{array}
\usepackage{subcaption}
\usepackage[ruled]{algorithm2e}
\usepackage[dvipsnames]{xcolor}
\usepackage{marvosym}
\usepackage[title]{appendix}
\usepackage[accsupp]{axessibility}  

\usepackage[pagebackref,breaklinks,colorlinks]{hyperref}

\usepackage[capitalize]{cleveref}
\crefname{section}{Sec.}{Secs.}
\Crefname{section}{Section}{Sections}
\Crefname{table}{Table}{Tables}
\crefname{table}{Tab.}{Tabs.}

\def\cvprPaperID{5345} 
\def\confName{CVPR}
\def\confYear{2023}

\begin{document}

\title{Multimodal Industrial Anomaly Detection via Hybrid Fusion}

\author{Yue Wang$^{1}$\footnotemark[1], Jinlong Peng$^{2}$\thanks{Equal contributions. This work was done when Yue Wang was a intern at Tencent Youtu Lab.}, Jiangning Zhang$^2$, Ran Yi$^{1}$\thanks{Corresponding author.}, Yabiao Wang$^2$, Chengjie Wang$^{2,1}$\\
$^1$Shanghai Jiao Tong University, Shanghai, China; $^2$Youtu Lab, Tencent\\
$^1${\tt\small \{imwangyue,ranyi\}@sjtu.edu.cn} $^2${\tt\small\{jeromepeng,vtzhang,caseywang,jasoncjwang\}@tencent.com}
}

\maketitle

\begin{abstract}
2D-based Industrial Anomaly Detection has been widely discussed, however, multimodal industrial anomaly detection based on 3D point clouds and RGB images still has many untouched fields. 
Existing multimodal industrial anomaly detection methods directly concatenate the multimodal features, which leads to a strong disturbance between features and harms the detection performance. 
In this paper, we propose {\bf \textit{Multi-3D-Memory}} ({\bf\textit{M3DM}}), a novel multimodal anomaly detection method with hybrid fusion scheme:
firstly, we design an unsupervised feature fusion with patch-wise contrastive learning to encourage the interaction of different modal features;
secondly, we use a decision layer fusion with multiple memory banks to avoid loss of information and additional novelty classifiers to make the final decision.
We further propose a point feature alignment operation to better align the point cloud and RGB features.
Extensive experiments show that our multimodal industrial anomaly detection model outperforms the state-of-the-art (SOTA) methods on both detection and segmentation precision on  MVTec-3D AD dataset.
Code at \href{https://github.com/nomewang/M3DM}{github.com/nomewang/M3DM}.

\end{abstract}

\section{Introduction}
\label{sec:intro}

\begin{figure}[t]
  \centering
  \includegraphics[width=0.86\linewidth]{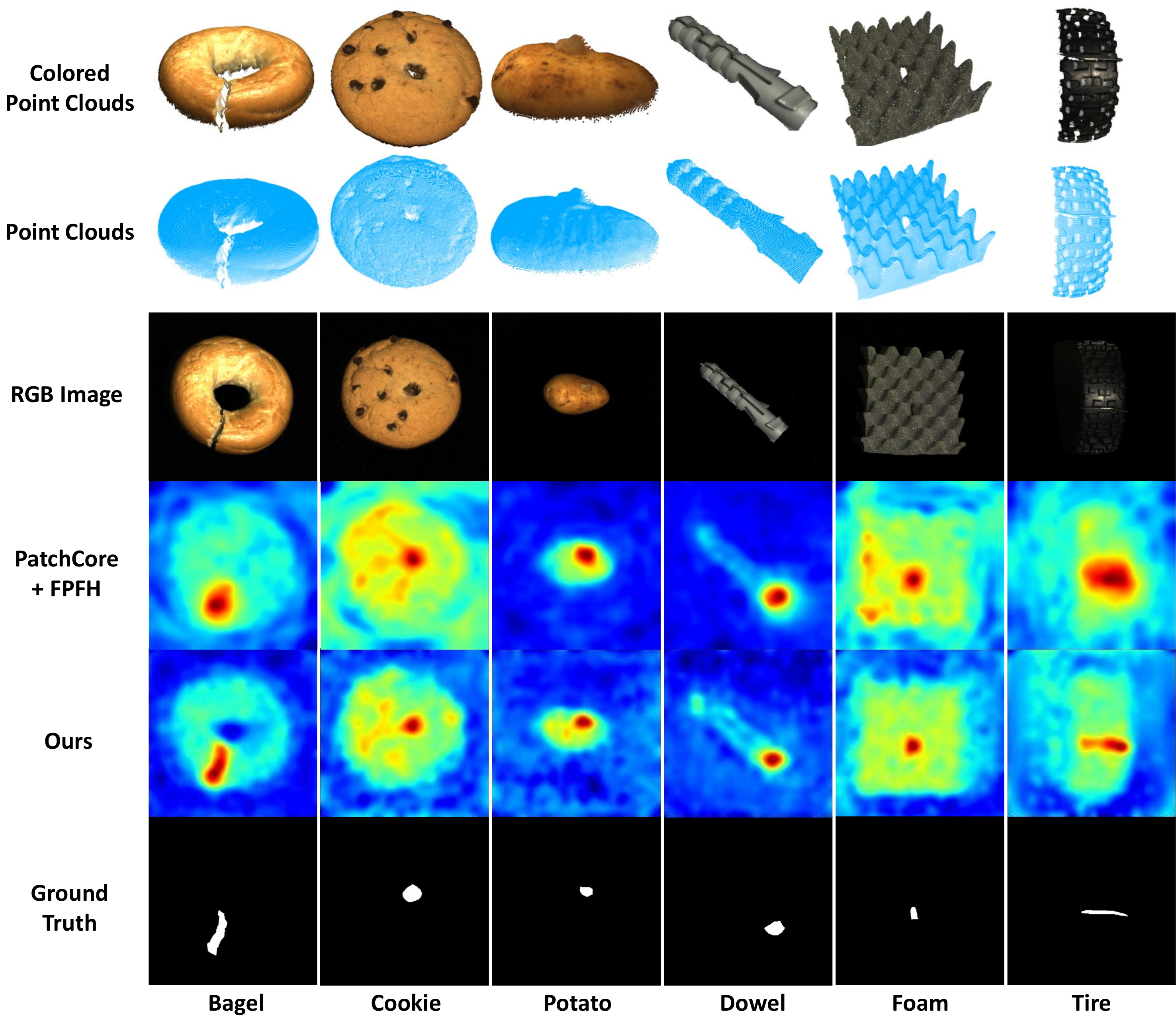}
  \vspace{-5pt}
   \caption{Illustrations of MVTec-3D AD dataset\cite{mvtec3dad}. The second and third rows are the input point cloud data and the RGB data. The fourth and fifth rows are prediction results, and according to the ground truth, our prediction has more accurate prediction results than the previous method.}
   \vspace{-10pt}
   \label{fig:teaser}
\end{figure}

Industrial anomaly detection aims to find the abnormal region of products and plays an important role in industrial quality inspection.
In industrial scenarios, it's easy to acquire a large number of normal examples, but defect examples are rare.
Current industrial anomaly detection methods are mostly unsupervised methods, i.e., only training on normal examples, and testing on detect examples only during inference.
Moreover, most existing industrial anomaly detection methods \cite{mvtec, draem, ReverseDistillation, patchcore} are based on 2D images.
However, in the quality inspection of industrial products, human inspectors utilize both the 3D shape and color characteristics to determine whether it is a defective product, where 3D shape information is important and essential for correct detection. 
As shown in \cref{fig:teaser}, for cookie and potato, it is hard to identify defects from the RGB image alone.
With the development of 3D sensors, recently MVTec-3D AD dataset \cite{mvtec3dad} (\cref{fig:teaser}) with both 2D images and 3D point cloud data has been released and facilitates the research on multimodal industrial anomaly detection.

The core idea for unsupervised anomaly detection is to find out the difference between normal representations and anomalies.
Current 2D industrial anomaly detection methods can be categorized into two categories:
(1) Reconstruction-based methods.
Image reconstruction tasks are widely used in anomaly detection methods \cite{mvtec, memorizing-ae, reconstruction, draem, ReverseDistillation, ocgan} to learn normal representation. 
Reconstruction-based methods are easy to implement for a single modal input (2D image or 3D point cloud).
But for multimodal inputs, it is hard to find a reconstruction target.
(2)  Pretrained feature extractor-based methods.
An intuitive way to utilize the feature extractor is to map the extracted feature to a normal distribution and find the out-of-distribution one as an anomaly.
Normalizing flow-based methods \cite{cflow,fastflow,ast} use an invertible transformation to directly construct normal distribution, and memory bank-based methods\cite{padim, patchcore} store some representative features to implicitly construct the feature distribution.
Compared with reconstruction-based methods, directly using a pretrained feature extractor does not involve the design of a multimodal reconstruction target and is a better choice for the multimodal task.
Besides that, current multimodal industrial anomaly detection methods \cite{3d-ads, ast} directly concatenate the features of the two modalities together.
However, when the feature dimension is high, the disturbance between multimodal features will be violent and cause performance reduction.

To address the above issues, we propose a novel multimodal anomaly detection scheme based on RGB images and 3D point cloud, named \textit{Multi-3D-Memory (M3DM)}.
Different from the existing methods that directly concatenate the features of the two modalities, we propose a hybrid fusion scheme to reduce the disturbance between multimodal features and encourage feature interaction.
We propose \textit{Unsupervised Feature Fusion (UFF)} to fuse multimodal features, which is trained using a patch-wise contrastive loss to learn the inherent relation between multimodal feature patches at the same position. 
To encourage the anomaly detection model to keep the single domain inference ability, we construct three memory banks separately for RGB, 3D and fused features.
For the final decision, we construct \textit{Decision Layer Fusion (DLF)} to consider all of the memory banks for anomaly detection and segmentation.

Anomaly detection needs features that contain both global and local information, 
where the local information helps detect small defects, 
and global information focuses on the relationship among all parts. 
Based on this observation, we utilize a Point Transformer\cite{point_transformer, pointmae} for the 3D feature and Vision Transformer\cite{vit, DINO} for the RGB feature. 
We further propose a \textit{Point Feature Alignment (PFA)} operation to better align the 3D and 2D features.

Our contributions are summarized as follows:
\begin{itemize}
  \item We propose M3DM, a novel multimodal industrial anomaly detection method with hybrid feature fusion,
  which outperforms the state-of-the-art detection and segmentation precision on MVTec-3D AD. 
  \item We propose Unsupervised Feature Fusion (UFF) with patch-wise contrastive loss to encourage interaction between multimodal features.
  \item We design Decision Layer Fusion (DLF) utilizing multiple memory banks for robust decision-making.
  \item We explore the feasibility of the Point Transformer in multimodal anomaly detection and propose Point Feature Alignment (PFA) operation to align the Point Transformer feature to a 2D plane for high-performance 3D anomaly detection.
\end{itemize}

\section{Related Works}
\label{sec:related}

\begin{figure*}[t]
  \centering
  \includegraphics[width=0.95\linewidth]{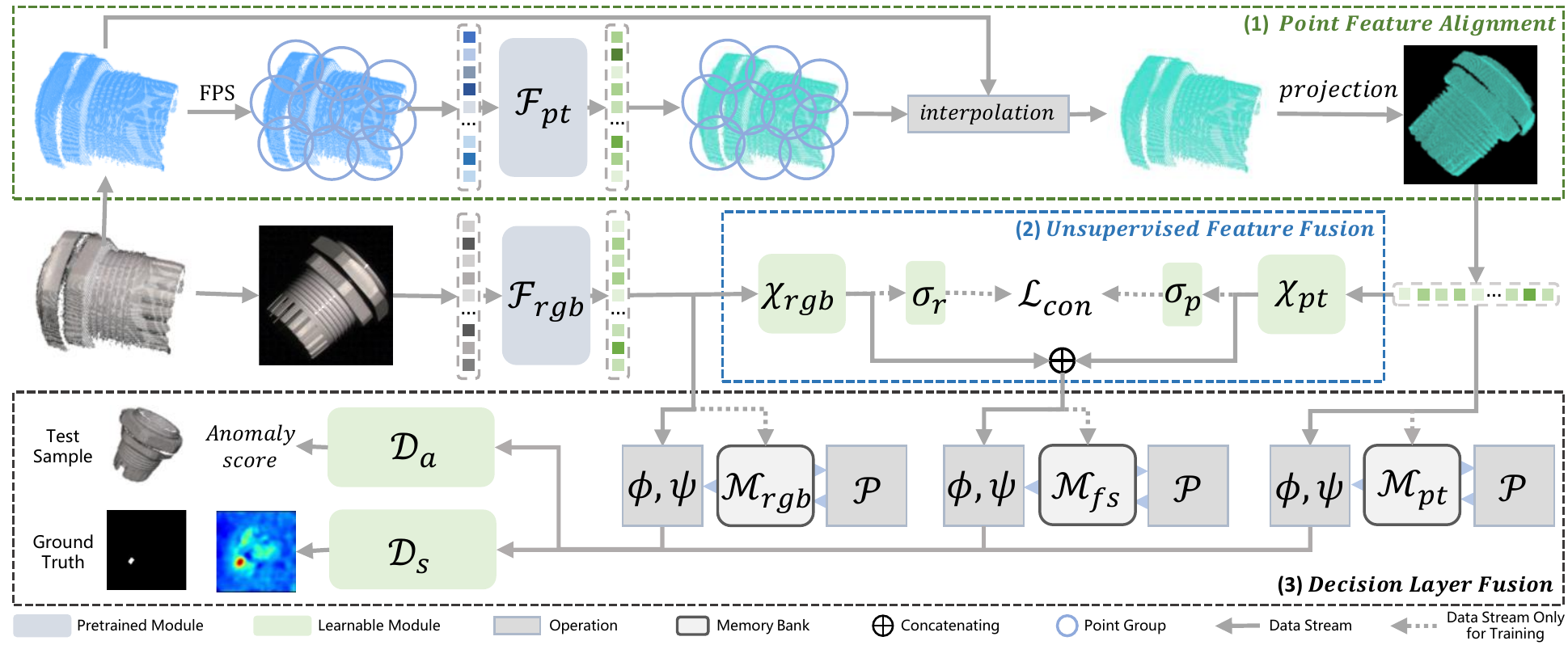}
   \caption{\textbf{The pipeline of  Multi-3D-Memory (M3DM).} Our M3DM contains three important parts: (1) {\bf \textit{ \color{OliveGreen}{Point Feature Alignment}}} (PFA) converts Point Group features to plane features with interpolation and project operation, $\text{FPS}$ is the farthest point sampling and $\mathcal F_{pt}$ is a pretrained Point Transformer; (2)
    {\bf \textit{ \color{RoyalBlue}{Unsupervised Feature Fusion}}} (UFF) fuses point feature and image feature together with a patch-wise contrastive loss $\mathcal L_{con}$, where $\mathcal F_{rgb}$ is a Vision Transformer, $\chi_{rgb},\chi_{pt}$ are MLP layers and $\sigma_r, \sigma_p$ are single fully connected layers; (3) {\bf \textit{Decision Layer Fusion}} (DLF) combines multimodal information with multiple memory banks and makes the final decision with 2 learnable modules $\mathcal D_a, \mathcal D_s$ for anomaly detection and segmentation, where $\mathcal{M}_{rgb}, \mathcal{M}_{fs}, \mathcal{M}_{pt}$ are memory banks, $\phi, \psi$ are score function for single memory bank detection and segmentation, and  $\mathcal{P}$ is the memory bank building algorithm.}
   \label{fig:pipeline}
\end{figure*}

For most anomaly detection methods, the core idea is to find out good representations of the normal data.
Traditional anomaly detection has developed several different roads.
Probabilistic-based methods use empirical cumulative distribution functions\cite{crosby1994detect, ecod} of normal samples to make decisions.
The position of representation space neighbors can also be used, and it can be done with several cluster methods, for example, k-NN\cite{knn, fastknn}, correlation integral\cite{loci} and histogram\cite{histogram}.
Outlier ensembles use a series of decision models to detect anomaly data, the most famous outlier ensembles method is Isolation Forest\cite{isolationforest}.
The linear model can also be used in anomaly detection, for example simply using the properties of principal component analysis\cite{PCA} or one-class support vector machine (OCSVM)\cite{OCSVM}.
The traditional machine learning method relies on less training data than deep learning, so we capture this advantage and design a decision layer fusion module based on OCSVM and stochastic gradient descent.

{\bf 2D Industrial Anomaly Detection}
Industrial anomaly detection is usually under an unsupervised setting.
The MVTec AD dataset is widely used for industrial anomaly detection\cite{mvtec} research, and it only contains good cases in the training dataset but contains both good and bad cases in the testing dataset.
Industrial anomaly detection needs to extract image features for decision, and the features can be used either implicitly or explicitly.
Implicit feature methods utilize some image reconstruction model, for example, auto-encoder\cite{mvtec, memorizing-ae, reconstruction} and generative adversarial network\cite{ocgan};
Reconstruction methods could not recover the anomaly region, and comparing the generated image and the original image could locate the anomaly and make decisions.
Some data augmentation methods\cite{draem} were proposed to improve the anomaly detection performance, in which researchers manually add some pseudo anomaly to normal samples and the training goal is to locate pseudo anomaly.
Explicit feature methods rely on the pretrained feature extractor, and additional detection modules learn to locate the abnormal area with the learned feature or representation.
Knowledge distillation methods\cite{ReverseDistillation} aim to learn a student network to reconstruct images or extract the feature, the difference between the teacher network and student network can represent the anomaly.
Normalizing flow\cite{cflow, fastflow} utilizes an invertible transformation to convert the image feature to Normal distribution, and the anomaly feature would fall on the edge of the distribution.
Actually, all of the above methods try to store feature information in the parameters of deep networks, recent work shows that simply using a memory bank\cite{patchcore} can get a total recall on anomaly detection.
There are many similarities between 2D and 3D anomaly detection, we extend the memory bank method to 3D and multimodal settings and get an impressive result.

{\bf 3D Industrial Anomaly Detection}
The first public 3D industrial anomaly detection dataset is MVTec-3D\cite{mvtec3dad} AD dataset, which contains both RGB information and point position information for the same instance. 
Inspired by medical anomaly detection voxel auto-encoder and generative adversarial network\cite{mvtec3dad} were first explored in 3D industrial anomaly detection, but those methods lost much spacial structure information and get a poor results. 
After that, a 3D student-teacher network\cite{3d-st} was proposed to focus on local point clouds geometry descriptor with extra data for pretraining.
Memory bank method\cite{3d-ads} has also been explored in 3D anomaly detection with geometry point feature and a simple feature concatenation.
Knowledge distillation method\cite{ast} further improved the pure RGB and multimodal anomaly detection results with Depth information.
Our method is based on memory banks, and in contrast to previous methods, we propose a novel pipeline to utilize a pretrained point transformer and a hybrid feature fusion scheme for more precise detection.

\section{Method}
\label{sec:method}

\subsection{Overview}
\label{subsec:Over}
Our Multi-3D-Memory (M3DM) method takes a 3D point cloud and an RGB image as inputs and conducts 3D anomaly detection and segmentation.
We propose a hybrid fusion scheme to promote cross-domain information interaction and maintain the original information of every single domain at the same time.
We utilize two pretrained feature extractors, DINO\cite{DINO} for RGB and PointMAE\cite{pointmae} for point clouds, to extract color and 3D representations respectively. 

As shown in \cref{fig:pipeline}, M3DM consists of three important parts: 
(1) Point Feature Alignment (PFA in \cref{subsec:PFA}): to solve the position information mismatch problem of the color feature and 3D feature, we propose Point Feature Alignment to align the 3D feature to 2D space, which helps simplify multimodal interaction and promotes detection performance.
(2) Unsupervised Feature Fusion (UFF in \cref{subsec:UFF}): since the interaction between multimodal features can generate new representations helpful to anomaly detection \cite{3d-ads, ast}, we propose an Unsupervised Feature Fusion module to help unify the distribution of multimodal features and learn the inherent connection between them.
(3) Decision Layer Fusion (DLF in \cref{subsec:DLF}): although UFF helps improve the detection performance, we found that information loss is unavoidable and propose Decision Layer Fusion to utilize multiple memory banks for the final decision.

\subsection{Point Feature Alignment}
\label{subsec:PFA}

{\bf Point Feature Extraction.}
We utilize a Point Transformer ($\mathcal F_{pt}$)~\cite{point_transformer} to extract the point clouds feature.
The input point cloud $p$ is a point position sequence with $N$ points. After the farthest point sampling (FPS)~\cite{pointnet++}, the point cloud is divided into $M$ groups, each with $S$ points.
Then the points in each group are encoded into a feature vector, and the $M$ vectors are input into the Point Transformer.
The output $g$ from the Point Transformer are $M$ point features, 
which are then organized as point feature groups: each group has a single point feature, which can be seen as the feature of the center point.

{\bf Point Feature Interpolation.}
Since after the farthest point sampling (FPS), the point center points are not evenly distributed in space, which leads to a unbalance density of point features.
We propose to interpolate the feature back to the original point cloud.
Given $M$ point features ${g_i}$ associated with $M$ group center points $c_i$, we use inverse distance weight to interpolate the feature to each point $p_j$ ($j \in \{ 1,2,...,N\}$) in the input point clouds.
The process can be described as:
\begin{equation}
\begin{aligned}
    p'_j = \sum_{i=1}^{M} \alpha_i g_i, \quad
    \alpha_i = \frac{ \frac{1}{\Vert c_i - p_j \Vert_2+\epsilon}} {\sum_{k=1}^M \sum_{t=1}^{N} \frac{1}{\Vert c_k - p_t \Vert_2+\epsilon}},
    \label{eq:interpolation}
\end{aligned}
\end{equation}
where $\epsilon$ is a fairly small constant to avoid $0$ denominator.

{\bf Point Feature Projection.}
After interpolation, we project $p'_j$ onto the 2D plane using the point coordinate and camera parameters, and we denote the projected points as $\hat{p}$.
We noticed that the point clouds could be sparse, if a 2D plane position doesn't match any point, we simply set the position as 0. 
We denote the projected feature map as $\{\hat{p}_{x,y}|(x,y)\in \mathbb{D}\}$ ($\mathbb{D}$ is the 2D plane region of the RGB image), which has the same size as the input RGB image.
Finally, we use an average pooling operation to get the patch feature on the 2D plane feature map.

\subsection{Unsupervised Feature Fusion}
\label{subsec:UFF}

\begin{figure}[t]
  \centering
  \includegraphics[width=0.95\linewidth]{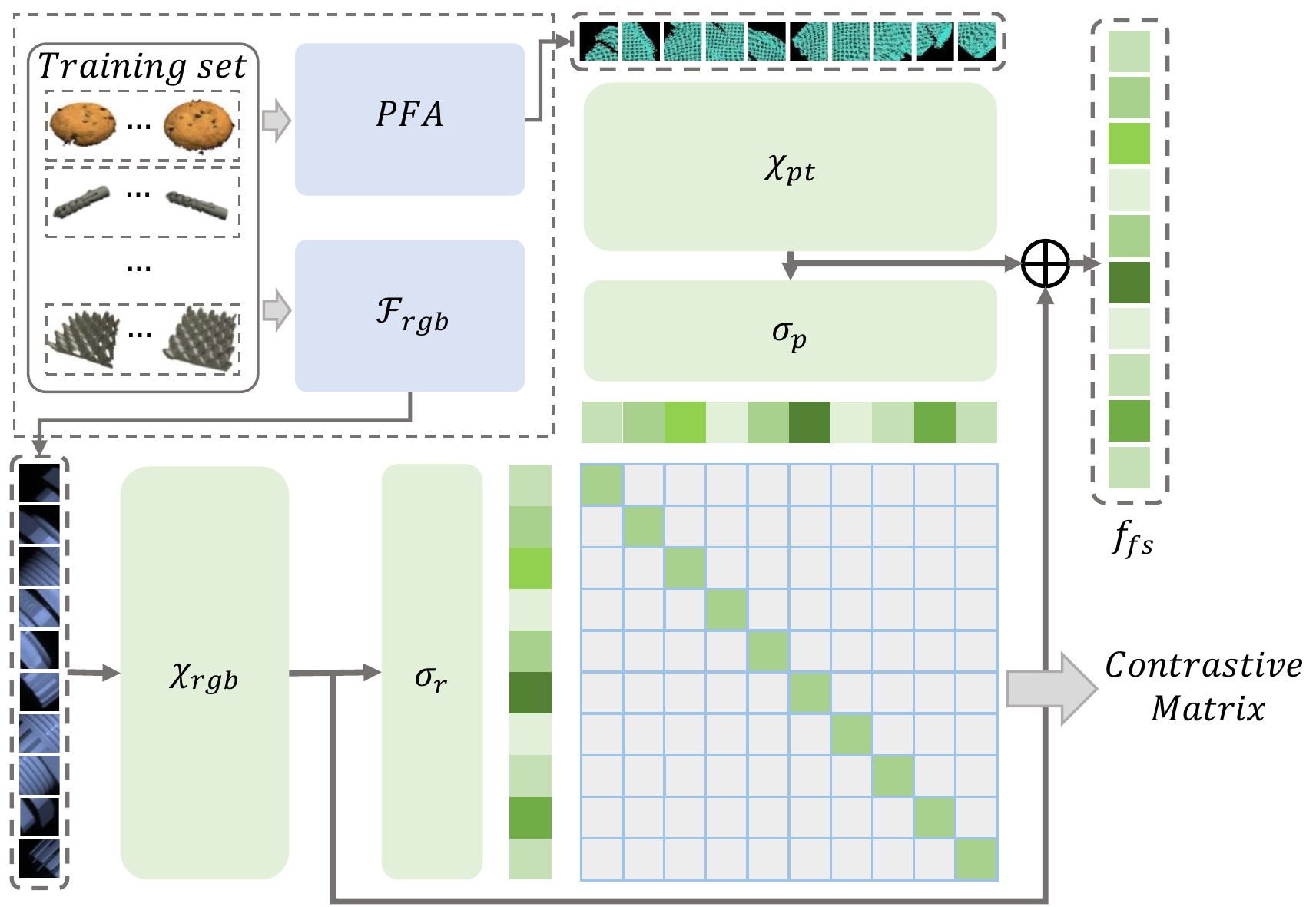}
   \caption{UFF architecture. UFF is a unified module trained with all training data of MVTec-3D AD. The patch-wise contrastive loss $\mathcal{L}_{con}$ encourages the multimodal patch features in the same position to have the most mutual information, i.e., the diagonal elements of the contrastive matrix have the biggest values.}
   \vspace{-5pt}
   \label{fig:UFF}
\end{figure}

The interaction between multimodal features can create new information that is helpful for industrial anomaly detection.
For example, in \cref{fig:teaser}, we need to combine both the black color and the shape depression to detect the hole on the cookie.
To learn the inherent relation between the two modalities that exists in training data, we design the Unsupervised Feature Fusion (UFF) module. 
We propose a patch-wise contrastive loss to train the feature fusion module:
given RGB features $f_{rgb}$ and point clouds feature $f_{pt}$, we aim to encourage the features from different modalities at the same position to have more corresponding information, while the features at different positions have less corresponding information.

We denote the features of a patch as $\{f^{(i,j)}_{rgb},f^{(i,j)}_{pt}\}$, where $i$ is the index of the training sample and $j$ is the index of the patch.
We conduct multilayer perceptron (MLP) layers $\{\chi_{rgb}, \chi_{pt} \}$ to extract interaction information between two modals and use fully connected layers $\{\sigma_{rgb}, \sigma_{pt} \}$ to map processed feature to query or key vectors. We denote the mapped features as $\{h^{(i,j)}_{rgb},h^{(i,j)}_{pt}\}$.
Then we adopt InfoNCE~\cite{infonce} loss for the contrastive learning:
\begin{equation}
\mathcal{L}_{con} = \frac{h_{rgb}^{(i,j)} \cdot h_{pt}^{(i,j)}{}^T}{\sum_{t=1}^{N_b} \sum_{k=1}^{N_p} h_{rgb}^{(t,k)} \cdot h_{pt}^{(t,k)}{}^T},
\end{equation}
where  $N_b$ is the batch size and $N_p$ is the nonzero patch number.
UFF is a unified module trained with all categories' training data of the MVTec-3D AD, and the architecture of UFF is shown in \cref{fig:UFF}.

During the inference stage, we concatenate the MLP layers outputs as a fused patch feature denoted as $f^{(i, j)}_{fs}$ :
\begin{equation}
    f^{(i,j)}_{fs} = \chi_{rgb}(f^{(i,j)}_{rgb}) \oplus \chi_{pt}(f^{(i,j)}_{pt}).
\end{equation}

\subsection{Decision Layer Fusion}
\label{subsec:DLF}

As shown in \cref{fig:teaser}, a part of industrial anomaly only appears in a single domain (e.g., the protruding part of potato), and the correspondence between multimodal features may not be extremely obvious.
Moreover, although Feature Fusion promotes the interaction between multimodal features, we still found that some information has been lost during the fusion process.

To solve the above problem, we propose to utilize multiple memory banks to store the original color feature, position feature and fusion feature.
We denote the three kind of memory banks as $\mathcal{M}_{rgb}, \mathcal{M}_{pt}, \mathcal{M}_{fs}$ respectively.
We refer PatchCore\cite{patchcore} to build these three memory banks, and during inference, each memory bank is used to predict an anomaly score and a segmentation map.
Then we use two learnable One-Class Support Vector Machines (OCSVM)\cite{OCSVM} $\mathcal D_a$ and $\mathcal D_s$ to make the final decision for both anomaly score $a$ and segmentation map ${\bf \textit{S}}$. 
We call the above process Decision Layer Fusion (DLF), which can be described as:
\begin{equation}
    a = \mathcal D_a (\phi(\mathcal{M}_{rgb},f_{rgb}), \phi(\mathcal{M}_{pt},f_{pt}), \phi(\mathcal{M}_{fs},f_{fs})),
    \label{eq:dlf_a}
\end{equation}
\begin{equation}
\small
    \textbf{\it S} = \mathcal D_s (\psi (\mathcal{M}_{rgb},f_{rgb}), \psi(\mathcal{M}_{pt},f_{pt}), \psi(\mathcal{M}_{fs},f_{fs})),
    \label{eq:dlf_s}
\end{equation}
where $\phi, \psi$ are the score functions introduced by \cite{patchcore}, which can be formulated as:
\begin{equation}
     \phi(\mathcal{M}, f) = \eta\Vert f^{(i,j), *} - m^*\Vert_2,
\end{equation}
\begin{equation}
     \psi(\mathcal{M}, f) = \{\min _{m\in \mathcal{M}} \Vert f^{(i,j)} - m\Vert_2 \Big{|} f^{(i,j)}\in f\},
\end{equation}
\begin{equation}
     f^{(i,j), *}, m^* = \arg \max _{f^{(i,j)}\in f} \arg \min _{m\in \mathcal{M}}\Vert  f^{(i,j)} - m\Vert_2,
\end{equation}
where $\mathcal{M} \in \{ \mathcal{M}_{rgb}, \mathcal{M}_{pt}, \mathcal{M}_{fs} \}$, $f \in \{f_{rgb}, f_{pt}, f_{fs}\}$ and $\eta$ is a re-weight parameter. 

We propose a two-stage training procedure: in the first stage we construct memory banks, and in the second stage we train the decision layer. 
The pseudo-code of DLF is shown as \cref{alg:mmb}.

\begin{algorithm}
\footnotesize
\caption{Decision Layer Fusion Training}\label{alg:mmb}
\KwIn {Memory bank building algorithm $\mathcal P$\cite{patchcore}, 
decision layer $\{\mathcal D_a,\mathcal D_s\}$, OCSVM loss function $\mathcal{L}_{oc}$\cite{OCSVM}}
\KwData {Training set features $\{\mathbb F_{rgb},\mathbb F_{pt},\mathbb F_{fs}\}$.}
\KwOut {Multimodal memory banks $\{ \mathcal{M}_{rgb}, \mathcal{M}_{pt}, \mathcal{M}_{fs}\}$, decision layer parameters $\{\Theta_{\mathcal D_a}, \Theta_{\mathcal D_s}\}$.}

\For{$ modal \in \{rgb, pt ,fs\}$}{
    \For{$f_{modal}\in \mathbb F_{modal}$}{
        $\mathcal{M}_{modal} \leftarrow f_{modal}$
    }
    $\mathcal{M}_{modal} \leftarrow \mathcal P(\mathcal{M}_{modal})$
}
\For{$f_{rgb}\in \mathbb F_{rgb}, f_{pt} \in  \mathbb F_{pt}, f_{fs} \in \mathbb F_{fs}$}{
{   \small
    $ \Theta_{\mathcal D_a} \stackrel{optim}{\longleftarrow} \mathcal{L}_{oc}(\mathcal D_a; \Theta_{\mathcal D_a})$ \\
    $ \Theta_{\mathcal D_s} \stackrel{optim}{\longleftarrow} \mathcal{L}_{oc}(\mathcal D_s; \Theta_{\mathcal D_s})$
    }
}
\end{algorithm}
\vspace{-15pt}

\section{Experiments}
\label{sec:experiments}

\subsection{Experimental Details}
\label{sec:experiments_detail}

{\bf Dataset.}
3D industrial anomaly detection is in the beginning stage. 
The MVTec-3D AD dataset is the first 3D industrial anomaly detection dataset.
Our experiments were performed on the MVTec-3D dataset.

MVTec-3D AD\cite{mvtec3dad} dataset consists of 10 categories, a total of 2656 training samples, and 1137 testing samples.
The 3D scans were acquired by an industrial sensor using structured light, and position information was stored in 3 channel tensors representing $x$, $y$ and $z$ coordinates.
Those 3 channel tensors can be single-mapped to the corresponding point clouds.
Additionally, the RGB information is recorded for each point.
Because all samples in the dataset are viewed from the same angle, the RGB information of each sample can be stored in a single image.
Totally, each sample of the MVTec-3D AD dataset contains a colored point cloud.

{\bf Data Preprocess.}
Different from 2D data, 3D ones are easier to remove the background information.
Following \cite{3d-ads}, we estimate the background plane with RANSAC\cite{RANSAC} and any point within 0.005 distance is removed.
At the same time, we set the corresponding pixel of removed points in the RGB image as 0.
This operation not only accelerates the 3D feature processing during training and inference but also reduces the background disturbance for anomaly detection.
Finally, we resize both the position tensor and the RGB image to $224 \times 224$ size, which is matched with the feature extractor input size.

{\bf Feature Extractors.}
We use 2 Transformer-based feature extractors to separately extract the RGB feature and point clouds feature:
1) For the RGB feature, we use a Vision Transformer (ViT)\cite{vit} to directly extract each patch feature, and in order to adapt to the anomaly detection task, we use a ViT-B/8 architecture for both efficiency and detection grain size;
For higher performance, we use the ViT-B/8 pretrained on ImageNet\cite{imagenet} with DINO\cite{DINO}, and this pretrained model recieves a $224 \times 224$ image and outputs totally $784$ patches feature for each image;
Since previous research shows that ViT concentrated on both global and local information on each layer, we use the output of the final layer with $768$ dimensions for anomaly detection. 
2) For the point cloud feature, we use a Point Transformer\cite{point_transformer, pointmae}, which is pretrained on ShapeNet\cite{shapenet} dataset, as our 3D feature extractor, and use the $\{3, 7, 11\}$ layer output as our 3D feature; Point Transformer firstly encodes point cloud to point groups which are similar with patches of ViT and each group has a center point for position and neighbor numbers for group size.
As described in \cref{subsec:PFA}, we separately test the setting $M=784, S=64$ and $M=1024, S=128$ for our experiments.
In the PFA operation, we separately pool the point feature to $28 \times 28$ and $56 \times 56$ for testing.

{\bf Learnable Module Details. }
M3DM has 2 learnable modules: the Unsupervised Feature Fusion module and the Decision Layer Fusion module. 1) For UFF, the $\chi_{rgb}, \chi_{pc}$ are 2 two-layer MLPs with $4\times$ hidden dimension as input feature;  We use AdamW optimizer, set learning rate as 0.003 with cosine warm-up in 250 steps and batch size as 16, we report the best anomaly detection results under 750 UFF training steps. 
2) For DLF, we use two linear OCSVMs with SGD optimizers, the learning rate is set as $1\times10^{-4}$ and train 1000 steps for each class.

\begin{table*}
  \centering
  \scriptsize
  \begin{tabular}{@{}cl|m{0.8cm}<{\centering}m{0.8cm}<{\centering}m{0.8cm}<{\centering}m{0.8cm}<{\centering}m{0.8cm}<{\centering}m{0.8cm}<{\centering}m{0.8cm}<{\centering}m{0.8cm}<{\centering}m{0.8cm}<{\centering}m{0.8cm}<{\centering}|m{0.8cm}<{\centering}}
    \toprule
    & Method & Bagel & Cable Gland & Carrot & Cookie & Dowel & Foam & Peach & Potato & Rope & Tire & Mean\\
    \midrule
     \multirow{10}*{\rotatebox{90}{3D}}&Depth GAN\cite{mvtec3dad} & 0.530 & 0.376 & 0.607 & 0.603 & 0.497 & 0.484 & 0.595 & 0.489 & 0.536 & 0.521 & 0.523\\
     ~&Depth AE\cite{mvtec3dad} & 0.468 & \underline{0.731} & 0.497 & 0.673 & 0.534 & 0.417 & 0.485 & 0.549 & 0.564 & 0.546 & 0.546\\
     ~&Depth VM\cite{mvtec3dad} & 0.510 & 0.542 & 0.469 & 0.576 & 0.609 & 0.699 & 0.450 & 0.419 & 0.668 & 0.520 & 0.546\\
     ~&Voxel GAN\cite{mvtec3dad} & 0.383 & 0.623 & 0.474 & 0.639 & 0.564 & 0.409 & 0.617 & 0.427 & 0.663 & 0.577 & 0.537\\
     ~&Voxel AE\cite{mvtec3dad} & 0.693 & 0.425 & 0.515&  0.790 & 0.494 & 0.558 & 0.537 & 0.484 & 0.639 & 0.583 & 0.571\\
     ~&Voxel VM\cite{mvtec3dad} & 0.750  & \bf{0.747} & 0.613 & 0.738 & 0.823 & 0.693 & 0.679 & 0.652 & 0.609 &  \underline{0.690} & 0.699\\
     ~&3D-ST\cite{3d-st} & 0.862& 0.484 & 0.832 & 0.894 & 0.848 & 0.663 & 0.763 & 0.687 & \bf{0.958} & 0.486 & 0.748 \\
     ~&FPFH\cite{3d-ads} & 0.825 & 0.551 & \underline{0.952} & 0.797 & \underline{0.883} & 0.582 & 0.758 & 0.889 & 0.929 & 0.653 & 0.782\\
     ~&AST\cite{ast} & \underline{0.881} & 0.576 & {\bf 0.965} & \underline{0.957} & 0.679 & {\bf 0.797} & {\bf 0.990} & \underline{0.915} & \underline{0.956} & 0.611 & \underline{0.833}\\
     ~&Ours & {\bf 0.941} & 0.651 & {\bf 0.965} & {\bf 0.969} & {\bf 0.905} & \underline{0.760} & \underline{0.880} & {\bf 0.974}& 0.926 & {\bf 0.765} & {\bf 0.874}\\
    \midrule
      \multirow{7}*{\rotatebox{90}{RGB}}&DifferNet\cite{differnet} & 0.859 & 0.703 & 0.643 & 0.435 & 0.797 & 0.790 & 0.787 & \underline{0.643} & 0.715 & 0.590 & 0.696\\
     &PADiM\cite{padim} & {\bf 0.975} & 0.775 & 0.698 & 0.582 & 0.959 & 0.663 & 0.858 & 0.535 & 0.832 & 0.760 & 0.764\\
     &PatchCore\cite{patchcore} & 0.876 & 0.880 & 0.791 & 0.682 & 0.912 & 0.701 & 0.695 & 0.618 & 0.841 & 0.702 & 0.770\\
     &STFPM\cite{STFPM} & 0.930 & 0.847 & \underline{0.890} & 0.575 & 0.947 & 0.766 & 0.710 & 0.598 & 0.965 & 0.701 & 0.793\\
     &CS-Flow\cite{cflow} & 0.941 & {\bf 0.930} & 0.827 & \underline{0.795} & {\bf 0.990} & \underline{0.886} & 0.731 & 0.471 &  \underline{0.986} & 0.745 & 0.830\\
     &AST\cite{ast} & \underline{ 0.947} & \underline{0.928} & 0.851 & {\bf 0.825} & \underline{0.981} & {\bf 0.951} & \underline{0.895} & 0.613 & {\bf 0.992}& {\bf 0.821} & {\bf 0.880}\\
     &Ours & 0.944 & 0.918 & {\bf 0.896} & 0.749 & 0.959 & 0.767 & {\bf 0.919} & {\bf 0.648} & 0.938 & \underline{0.767} & \underline{0.850}\\
    \midrule
     \multirow{10}*{\rotatebox{90}{RGB + 3D}}&Depth GAN\cite{mvtec3dad} & 0.538 & 0.372& 0.580& 0.603& 0.430& 0.534& 0.642& 0.601& 0.443& 0.577& 0.532\\
     &Depth AE\cite{mvtec3dad} & 0.648 & 0.502 & 0.650 & 0.488& 0.805 & 0.522& 0.712 & 0.529 & 0.540 & 0.552 & 0.595\\
     &Depth VM\cite{mvtec3dad} & 0.513& 0.551& 0.477 & 0.581 & 0.617 & 0.716 & 0.450 & 0.421& 0.598& 0.623& 0.555\\
     &Voxel GAN\cite{mvtec3dad} & 0.680& 0.324& 0.565 & 0.399& 0.497& 0.482& 0.566& 0.579& 0.601& 0.482& 0.517\\
     &Voxel AE\cite{mvtec3dad} & 0.510& 0.540 & 0.384& 0.693& 0.446& 0.632& 0.550& 0.494& 0.721& 0.413& 0.538\\
     &Voxel VM\cite{mvtec3dad} & 0.553 & 0.772& 0.484& 0.701& 0.751& 0.578& 0.480& 0.466& 0.689& 0.611& 0.609\\
     &PatchCore + FPFH\cite{3d-ads} & 0.918 & 0.748 & 0.967 & 0.883 & \underline{0.932} & 0.582 & 0.896 & \underline{0.912} & 0.921 & {\bf 0.886} & 0.865\\
     &AST\cite{ast}  & \underline {0.983} & \underline{0.873} & \bf{0.976} & \underline{0.971} & \underline{0.932} & \underline{0.885} & {\bf 0.974} & \bf{0.981} & {\bf 1.000} & 0.797 & \underline{0.937} \\
     &Ours & {\bf 0.994} & {\bf 0.909} & \underline{0.972} & {\bf 0.976} & {\bf 0.960} & {\bf 0.942} & \underline{0.973} & 0.899& \underline{0.972} & \underline{0.850} & {\bf 0.945}\\
    \bottomrule
  \end{tabular}
  \vspace{-5pt}
  \caption{I-AUROC score for anomaly detection of all categories of MVTec-3D AD. Our method clearly outperforms other methods in 3D and 3D + RGB setting; For pure 3D setting, our method reaches 0.874 mean I-AUROC score, and for 3D + RGB setting, we get 0.945 mean I-AUROC score. The results of baselines are from the ~\cite{mvtec3dad, 3d-ads, ast, benchmarking}.}
  \vspace{-10pt}
  \label{tab:iaucroc}
\end{table*}

{\bf Evaluation Metrics.}
All evaluation metrics are exactly the same as in \cite{mvtec3dad}.
We evaluate the image-level anomaly detection performance with the area under the receiver operator curve (I-AUROC), and higher I-AUROC means better image-level anomaly detection performance. 
For segmentation evaluation, we use the per-region overlap (AUPRO) metric, 
which is defined as the average relative overlap of the binary prediction with each connected component of the ground truth.
Similar to I-AUROC, the receiver operator curve of pixel level predictions can be used to calculate P-AUROC for evaluating the segmentation performance.

\subsection{Anomaly Detection on MVTec-3D AD}
\label{subsec:ad_results}

We compare our method with several 3D, RGB and RGB + 3D multimodal methods on MVTec-3D, \cref{tab:iaucroc} shows the anomaly detection results record with I-AUROC, \cref{tab:aupro} shows the segmentation results report with AUPRO and we report the P-AUROC in supplementary materials.
1) On pure 3D anomaly detection we get the highest I-AUROC and outperform AST\cite{ast} 4.1\%, which shows our method has much better detection performance than the previous method, and with our PFA, the Point Transformer is the better 3D feature extractor for this task; for segmentation, we get the second best result with AUPRO as 0.906, since our 3D domain segmentation is based on the point cloud, we find there is a bias between point clouds and ground truth label and discuss this problem in \cref{subsec:discuss}.
2) On RGB anomaly detection, the difference between our method and Patchcore\cite{patchcore} is that we use a Transformer based feature extractor instead of a Wide-ResNet one and remove the pooling operation before building the memory bank; Our I-AUROC in RGB domain is 8.0\% higher than the original PatchCore results and get the highest AUPRO score for segmentation, which is 7.6\% higher than the second best one.
3) On RGB + 3D multimodel anomaly detection, our method gets the best results on both I-AUROC and AUPRO scores, we get 0.8\% better I-AUROC than the AST and 0.5\% better AUPRO than the PatchCore + FPFH \cite{3d-ads}; These results are contributed by our fusion strategy and the high-performance 3D anomaly detection results. The previous method couldn't have great detection and segmentation performance at the same time, as shown in \cref{tab:p-aucroc}. Since the AST\cite{ast} didn't report the AUPRO results, we compare the segmentation performance with P-AUROC here.
Although PatchCore + FPFH method gets a high P-AUROC score, its I-AUROC is much lower than the other two.
Besides, AST gets a worse P-AUROC score than the other two methods, which means the AST is weak in locating anomalies.

\begin{table*}
  \centering
  \scriptsize
  \begin{tabular}{@{}cl|m{0.8cm}<{\centering}m{0.8cm}<{\centering}m{0.8cm}<{\centering}m{0.8cm}<{\centering}m{0.8cm}<{\centering}m{0.8cm}<{\centering}m{0.8cm}<{\centering}m{0.8cm}<{\centering}m{0.8cm}<{\centering}m{0.8cm}<{\centering}|m{0.8cm}<{\centering}}
    \toprule
    & Method & Bagel & Cable Gland & Carrot & Cookie & Dowel & Foam & Peach & Potato & Rope & Tire & Mean\\
    \midrule
     \multirow{8}*{\rotatebox{90}{3D}}&Depth GAN\cite{mvtec3dad} & 0.111& 0.072& 0.212& 0.174 & 0.160 & 0.128 & 0.003 & 0.042 & 0.446 & 0.075 & 0.143\\
     ~&Depth AE\cite{mvtec3dad} & 0.147 & 0.069 & 0.293 & 0.217 & 0.207 & 0.181 & 0.164 & 0.066 & 0.545 & 0.142 & 0.203 \\
     ~&Depth VM\cite{mvtec3dad} & 0.280 & 0.374 & 0.243 & 0.526 & 0.485 & 0.314 & 0.199 & 0.388 & 0.543 & 0.385 & 0.374\\
     ~& Voxel GAN\cite{mvtec3dad} & 0.440 & 0.453 & 0.875 & 0.755 & 0.782 & 0.378 & 0.392 & 0.639 & 0.775 & 0.389 & 0.583 \\
     ~&Voxel AE\cite{mvtec3dad} & 0.260 & 0.341 & 0.581 & 0.351 & 0.502 & 0.234 & 0.351 & 0.658 & 0.015 & 0.185 & 0.348 \\
     ~&Voxel VM\cite{mvtec3dad} & 0.453 & 0.343 & 0.521 & 0.697 & 0.680 & 0.284 & 0.349 & 0.634 & 0.616 & 0.346 & 0.492\\
     ~&3D-ST\cite{3d-st} & \underline{0.950} & 0.483 & {\bf 0.986} & {\bf 0.921} & {\bf 0.905} & 0.632 & 0.945 & {\bf 0.988} & {\bf 0.976} & 0.542 & 0.833\\
     ~&FPFH\cite{3d-ads} & {\bf 0.973} & {\bf 0.879} & {\underline{0.982}} & {\underline{ 0.906}} & {\underline{0.892}} & \underline{0.735} & {\bf 0.977} & {\underline{0.982}} & {\underline{0.956}} & {\bf 0.961} & {\bf 0.924}\\
     ~&Ours & 0.943 & \underline{0.818} & 0.977 & 0.882 & 0.881 & {\bf 0.743} & \underline{0.958} & 0.974 & 0.95 & \underline{0.929} & \underline{0.906} \\
    \midrule
     \multirow{4}*{\rotatebox{90}{RGB}}& CFlow\cite{cflow} & 0.855 & 0.919 & \underline{0.958} & 0.867 & \bf 0.969 & 0.500 & 0.889 & 0.935 & 0.904 & 0.919 & 0.871 \\
     & PatchCore\cite{patchcore} & 0.901 & \underline{0.949} & 0.928 & 0.877 & 0.892 & 0.563 & 0.904 & 0.932 & 0.908 & 0.906 & 0.876\\
     &  PADiM\cite{padim} & \bf 0.980 &  0.944 & 0.945 & \underline{0.925} & \underline{0.961} & \underline{0.792} & \underline{0.966} & \underline{0.940} & \underline{0.937} & \underline{0.912} & \underline{0.930}  \\
     & Ours & \underline{0.952}& {\bf 0.972} & {\bf 0.973} & {\bf 0.891} & 0.932 & {\bf 0.843} & {\bf 0.97} & {\bf 0.956} & {\bf 0.968} & {\bf 0.966} & {\bf 0.942} \\
    \midrule
     \multirow{9}*{\rotatebox{90}{RGB + 3D}}&Depth GAN\cite{mvtec3dad} & 0.421 & 0.422& 0.778 & 0.696 & 0.494 & 0.252 & 0.285 & 0.362 & 0.402 & 0.631 & 0.474 \\
     &Depth AE\cite{mvtec3dad} & 0.432 & 0.158 & 0.808 & 0.491 & 0.841 & 0.406 & 0.262 & 0.216 & 0.716 & 0.478 & 0.481 \\
     &Depth VM\cite{mvtec3dad} & 0.388 & 0.321 & 0.194 & 0.570 & 0.408 & 0.282 & 0.244 & 0.349 & 0.268 & 0.331 & 0.335\\
     &Voxel GAN\cite{mvtec3dad} & 0.664 & 0.620 & 0.766 & 0.740 & 0.783 & 0.332 & 0.582 & 0.790 & 0.633 & 0.483 & 0.639 \\
     &Voxel AE\cite{mvtec3dad} & 0.467 & 0.750 & 0.808 & 0.550 & 0.765 & 0.473 & 0.721 & 0.918 & 0.019 & 0.170 & 0.564\\
     &Voxel VM\cite{mvtec3dad} & 0.510 & 0.331 & 0.413 & 0.715 & 0.680 & 0.279 & 0.300 & 0.507 & 0.611 & 0.366 & 0.471\\
     &PatchCore + FPFH\cite{3d-ads} & {\bf 0.976} & \underline{0.969} & \bf{0.979} & {\bf 0.973} & \underline{0.933} & \underline{0.888} & \underline{0.975} & \bf{0.981} & \underline{0.950} & \underline{0.971} & \underline{0.959} \\
     &Ours & \underline{0.970} & {\bf 0.971} & \bf{0.979} & \underline{0.950} & {\bf 0.941} & {\bf 0.932} & {\bf 0.977} & \underline{0.971}  & \bf{0.971} & {\bf 0.975} & {\bf 0.964}\\
    \bottomrule
  \end{tabular}
  \vspace{-5pt}
  \caption{AUPRO score for anomaly segmentation of all categories of MVTec-3D. Our method outperforms other methods on RGB and RGB + 3D settings. for the RGB setting, our method reaches 0.942 mean AUPRO score, and for the RGB + 3D setting, our method reaches 0.964 mean AUPRO score. The results of baselines are from the ~\cite{mvtec3dad, 3d-ads, benchmarking}.}
  \vspace{-10pt}
  \label{tab:aupro}
\end{table*}

\begin{table}
  \centering
  \scriptsize
  \begin{tabular}{@{}l|cc}
    \toprule
    Method  & I-AUROC & P-AUROC\\
    \midrule
     PatchCore + FPFH \cite{3d-ads} & 0.865 & 0.992 \\
     AST\cite{ast} & 0.937 & 0.976\\
     Ours & {\bf 0.945}& {\bf 0.992}\\
    \bottomrule
  \end{tabular}
  \vspace{-5pt}
  \caption{Mean I-AUROC and P-AUROC score for anomaly detection of all categories of MVTec-3D. Our method performance well on both anomaly detection and segmentation.}
  \vspace{-5pt}
  \label{tab:p-aucroc}
\end{table}

\subsection{Ablation Study}
\label{subsec:ablation}

We conduct an ablation study on a multimodal setting, and to demonstrate our contributions to multimodal fusion, we analyze our method in \cref{tab:ablation} with the following settings:
1) Only Point Clouds ($\mathcal{M}_{pt}$) information; 
2) Only RGB ($\mathcal{M}_{rgb}$) information; 
3) Single memory bank ($\mathcal{M}_{fs}$) directly concatenating Point Transformer feature and RGB feature together;
4) Single memory bank ($\mathcal{M}_{fs}$) using UFF to fuse multimodal features;
5) Building two memory banks ($\mathcal{M}_{rgb},\mathcal{M}_{pt}$) separately and directly adding the scores together; 
6) Building two memory banks separately ($\mathcal{M}_{rgb},\mathcal{M}_{pt}$) and using DLF for the final result;
7) Building three memory banks ($\mathcal{M}_{rgb},\mathcal{M}_{pt},\mathcal{M}_{fs}$) (Ours).
Comparing row 3 and 4 in \cref{tab:ablation}, we can find that adding UFF greatly improves the results on all three metrics (I-AUROC 4.1\% $\uparrow $, AURPO 1.2\% $\uparrow $ and P-AUROC 0.3\% $\uparrow $), which shows UFF plays an important role for multimodal interaction and helps unify the feature distribution;
Compare row 5 and 6, we demonstrate that DLF model helps improve both anomaly detection and segmentation performance (I-AUROC 0.3\% $\uparrow $, AURPO 0.6\% $\uparrow $ and P-AUROC 0.3\% $\uparrow $). 
Our full set is shown in row 7 in \cref{tab:ablation}, and compared with row 6 we have 1.3\% I-AUROC and 0.5\% AUPRO improvement, which further demonstrates that UFF activates the interaction between two modals and creates a new feature for anomaly detection.   

\begin{table}
  \centering
  \scriptsize
  \begin{tabular}{@{}l|cccc}
    \toprule
   Method  & Memory bank &  I-AUROC & AUPRO & P-AUROC\\
    \midrule
    Only PC &$\mathcal{M}_{pt}$&0.874 & 0.906 &  0.970 \\
    Only RGB  &$\mathcal{M}_{rgb}$& 0.850 & 0.942 & 0.987 \\
    \midrule
    w/o UFF &$\mathcal{M}_{fs}$& 0.857  & 0.944& 0.987 \\
    w/ UFF &$\mathcal{M}_{fs}$ & 0.898  & 0.956 & 0.990 \\
    \midrule
     w/o DLF  & $\mathcal{M}_{rgb},\mathcal{M}_{pt}$ & 0.929 & 0.953 & 0.987 \\
     w/ DLF   & $\mathcal{M}_{rgb},\mathcal{M}_{pt}$ & 0.932 & 0.959 & 0.990 \\
     \midrule
     Ours  & $\mathcal{M}_{rgb},\mathcal{M}_{pt},\mathcal{M}_{fs}$ & {\bf 0.945} & {\bf 0.964}& {\bf 0.992}\\
    \bottomrule
  \end{tabular}
  \vspace{-5pt}
  \caption{Ablation study on fusion block. M is the number of memory banks used. Compared with directly concatenating feature, with UFF, the single memory bank method get better performance. With DLF, the anomaly detection and segmentation performance gets great improvement.}
  \vspace{-10pt}
  \label{tab:ablation}
\end{table}

\subsection{Analysis of PFA Hyper-parameter}
\label{subsec:PointTransformerSetting}

Since we are the first to use Point Transformer for 3D anomaly detection, we conduct a series of exploring experiments on the Point Transformer setting.
1) We first explore two important hyper-parameters of Point Transformer: the number of groups and the groups' size during farthest point sampling.
The number of groups decides how many features will be extracted by the Point Transformer and the groups' size is equal to the concept of the receptive field.
As shown in \cref{tab:pt-setting}, the model with 1,024 point groups and 128 points per group performs better in this task,
which we think is because more feature vectors help the model find refined representation and a suitable neighbor number would give more local position information.
2) To verify the PFA operation, we conduct another 3D anomaly detection experiment with the original point groups feature: a point group can be seen as a {\it patch}, and the memory bank store point groups feature here; 
The detection method is as same as the patch-based one, and to get the segmentation predictions, we first project point group feature to a 2D plane and use an inverse distance interpolation to get every pixel value; 
As shown in \cref{tab:pt-setting} row 2, the group-based method has better performance than its peer in row 4, however, when the patch-size gets smaller, the PFA-based method gets the best result on three metrics.

\begin{table}
  \centering
  \scriptsize
  \begin{tabular}{@{}ccc|ccc}
    \toprule
     S.G & N.G & Sampling &I-AUROC & AUPRO & P-AUROC\\
    \midrule
    64 & 784  & point group & 0.793 & 0.813 & 0.922 \\
    128 & 1024 & point group & 0.841 & 0.896 & 0.960 \\
    64 & 784 &$28\times28$ patches& 0.805 & 0.879  & 0.963 \\
    128 & 1024 &$28\times28$ patches  & 0.819 & 0.896 &  0.967\\
    128 & 1024 &$56\times56$ patches  & {\bf 0.874} & {\bf 0.906} &  {\bf 0.970}  \\
    \bottomrule
  \end{tabular}
  \vspace{-5pt}
  \caption{Exploring Point Transformer setting on the pure 3D setting. S.G means the point number per group, and N.G means the total number of point groups.  We get the best performance with 1024 point groups per sample and each point group contains 128 points; Compared with directly calculating anomaly and segmentation scores on point groups, the method based on a 2D plane patch needs a small patch size towards high performance.}
  \label{tab:pt-setting}
   \vspace{-5pt}
\end{table}

\subsection{Analysis of Multimodal Feature distribution}
\label{subsec:distribution}

\begin{figure}[t]
    \subfloat[Statistic distribution.]{
        \begin{minipage}[t]{0.46\linewidth}
            \centering
            \includegraphics[width=0.9\linewidth]{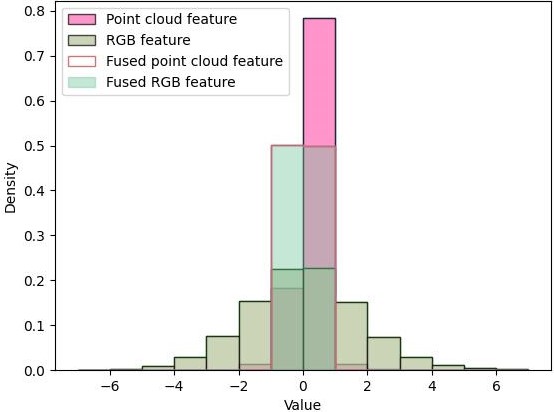}
            \label{fig:hist}
        \end{minipage}
    }
    \hfill
    \subfloat[T-SNE distribution.]{
        \begin{minipage}[t]{0.46\linewidth}
            \centering
            \includegraphics[width=0.9\linewidth]{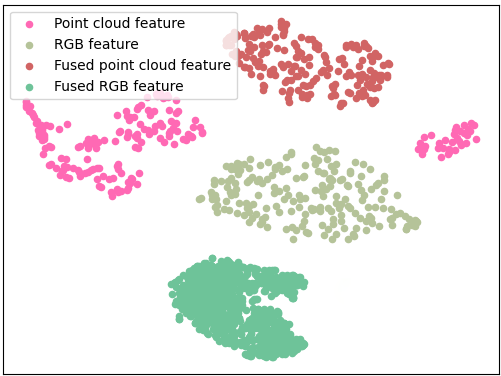}
            \label{fig:tsne}
        \end{minipage}
    }
    \vspace{-5pt}
     \caption{Distribution of bagel multimodal features. The fused point feature has a smaller variance in distribution and has a closer distribution with the fused RGB feature.}
      \vspace{-5pt}
     \label{fig:distri}
\end{figure}

We visualize the feature distribution with histogram and t-SNE\cite{t-SNE}. 
The original point cloud features have two disconnected regions in the t-SNE map (\cref{fig:tsne}), and it is caused by the pooling operation on the edge between the non-point region and the point cloud region.
The two fused features have similar distribution (in the \cref{fig:hist}), and these properties make the concatenated feature more suitable for memory bank building and feature distance calculation.
The original features have a more complex distribution, which is helpful for single-domain anomaly detection.
Our hybrid fusion scheme integrates the advantage of both original features and fused features and thus has a better performance than the single memory bank method.

\subsection{Few-shot Anomaly detection}
\label{subsec:few-shot}

We evaluate our method on Few-shot settings, and the results are illustrated in \cref{tab:few-shot}.
We randomly select 10 and 5 images from each category as training data and test the few-shot model on the full testing dataset.
We find that our method in a 10-shot or 5-shot setting still has a better segmentation performance than some non-few-shot methods.

\begin{table}
  \centering
  \scriptsize
  \begin{tabular}{@{}l|ccc}
    \toprule
    Method  & I-AUROC & AUPRO & P-AUROC\\
    \midrule
    5-shot & 0.796 & 0.927 & 0.981 \\
    10-shot & 0.821 & 0.939 & 0.985 \\
    50-shot & 0.903 & 0.953 & 0.988 \\
    Full dataset  & 0.945 & 0.964 &  0.992  \\
    \bottomrule
  \end{tabular}
  \vspace{-5pt}
  \caption{Few-shot setting results. On 10-shot or 5-shot setting, our method still has good segmentation performance and outperforms most Non-few-shot methods.}
  \vspace{-10pt}
  \label{tab:few-shot}
\end{table}

\subsection{Discussion about the MVTec-3D AD}
\label{subsec:discuss}

\begin{figure}[t]
  \centering
  \includegraphics[width=0.9\linewidth]{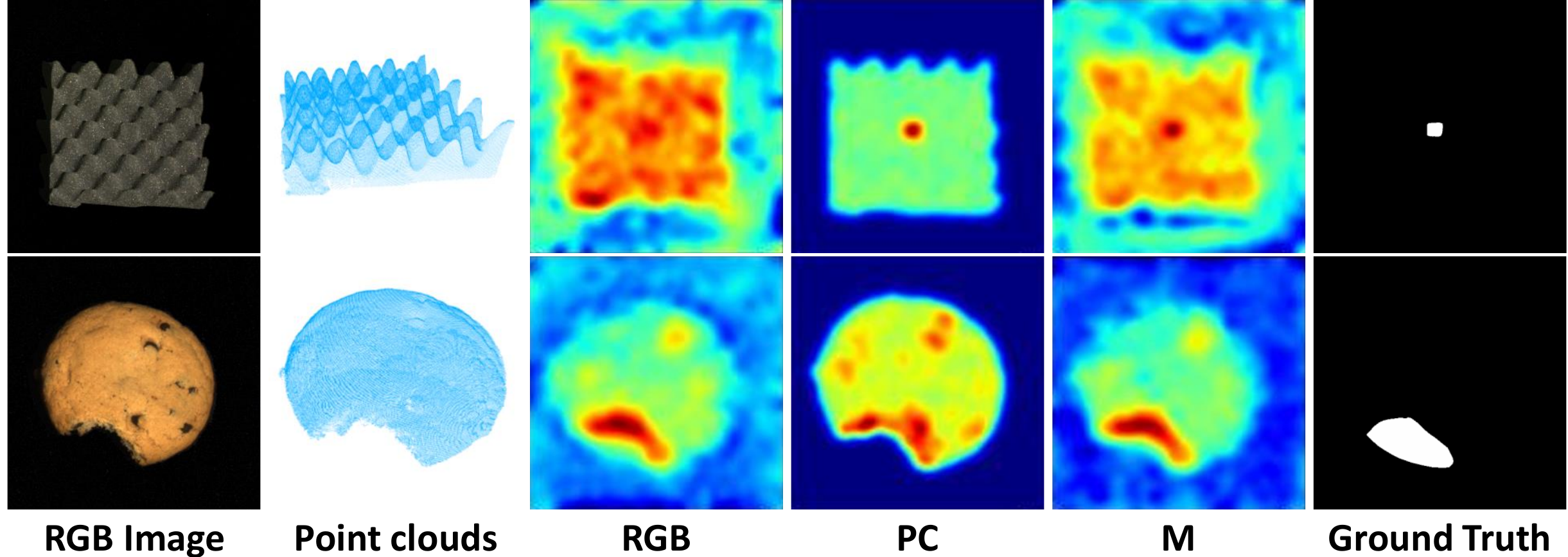}
  \vspace{-5pt}
   \caption{
      Properties of the MVTec-3D AD. PC is short for point cloud prediction, and M is short for multimodal prediction. With 3D information on the point cloud, the anomaly is accurately located in row 1. The label bias causes inaccurate segmentation, for the model hard to focus on the missing area.
   }
   \vspace{-10pt}
   \label{fig:discuss}
\end{figure}

In this section, we discuss some properties of the MVTec-3D AD dataset.
1) The 3D information helps detect more kinds of anomalies.
In the first row of \cref{fig:discuss}, we can find that the model fails to detect the anomaly with RGB information, but with the point cloud, the anomaly is accurately predicted.
This indicates that 3D information indeed plays an important role in this dataset. 
2) The label bias will cause inaccurate segmentation.
As shown in \cref{fig:discuss}, the cookie of row 2 has some cuts, and the anomaly label is annotated on the missing area.
However, for the pure point clouds method, the non-point region will not be reported, instead, the cut edge will be reported as an anomaly region, the phenomenon can be seen in the PC prediction of cookie in the \cref{fig:discuss}.
Because of this kind of bias between 3D point clouds and the 2D ground truth, the 3D version has a lower AUPRO score than the RGB one in \cref{tab:aupro}.
with the RGB information, the missing region will be more correctly reported as an anomaly.
Although we successfully predict more anomaly areas with multimodal data, there is still a gap between the prediction map and the ground truth.
We will focus on resolving this problem in future research.

\section{Conclusion}
\label{sec:conclusion}

In this paper, we propose a multimodal industrial anomaly detection method with point clouds and RGB images.
Our method is based on multiple memory banks and we propose a hybrid feature fusion scheme to process the multimodal data.
In detail, we propose a patch-wise contrastive loss-based Unsupervised Feature Fusion to promote multimodal interaction and unify the distribution,
and then we propose Decision Layer Fusion to fuse multiple memory bank outputs.
Moreover, we utilize pretrained Point Transformer and Vision Transformer as our feature extractors, and to align the above two feature extractors to the same spatial position, we propose Point Feature Alignment to convert 3D features to a 2D plane.
Our method outperforms the SOTA results on MVTec-3D AD datasets and we hope our work be helpful for further research.

\noindent{\bf Acknowledgements.}
This work was supported by  National Natural Science Foundation of China (72192821, 61972157, 62272447), Shanghai Municipal Science and Technology Major Project (2021SHZDZX0102), Shanghai Science and Technology Commission (21511101200), Shanghai Sailing Program (22YF1420300, 23YF1410500), CCF-Tencent Open Research Fund (RAGR20220121) and Young Elite Scientists Sponsorship Program by CAST (2022QNRC001).

{\small
\bibliographystyle{ieee_fullname}
\bibliography{egbib}
}

\clearpage
\appendix
\begin{subappendices}
\setcounter{table}{0}
\setcounter{figure}{0}
\renewcommand\thesection{\Alph{section}}
\renewcommand\thetable{\Roman{table}}
\renewcommand\thefigure{\Roman{figure}}

\section*{Overview}
\label{sec:overview}
This supplementary material includes:
\begin{itemize}
\item The implementation details about the hardware and software packages (\cref{sec:implementation});
\item P-AUROC score for anomaly segmentation (\cref{sec:pauroc});
\item Detailed score for ablation study of all categories of MVTec-3D AD (\cref{sec:ablation});
\item Detailed score for Point Transformer setting of all categories of MVTec-3D AD (\cref{sec:transformer});
\item Detailed score for few-shot setting of all categories of MVTec-3D AD (\cref{sec:few-shot});
\item Discussion about Backbone Choices (\cref{sec:backbone});
\item The results of all categories of Eyecandies (\cref{sec:eyecandies});
\item The visualization results of all categories of MVTec-3D AD (\cref{sec:visual}).
\end{itemize}

\section{Implementation Details}
\label{sec:implementation}

We implement M3DM with Pytorch\footnote{https://pytorch.org/} and Scikit-Learn package\footnote{https://scikit-learn.org/}. 
The feature extractors and memory banks algorithm are based on Pytorch and we use the Scikit-Learn package for OCSVM\cite{OCSVM}.
The AUROC calculation also relies on Scikit-Learn package.
All experiments are run on a single Nvidia Tesla V100 and cost at most 50 GB of memories for the full setting.

\section{P-AUROC for Segmentation}
\label{sec:pauroc}
In the main paper, we report the AUPRO score for anomaly segmentation.
In this section, we report the P-AUROC score to further verify the segmentation performance of our method, as shown in \cref{tab:pauroc}. 
We mainly compare our results with FPFH \cite{3d-ads}, PatchCore \cite{patchcore} and AST \cite{ast}\footnote{Since AST \cite{ast} only provided the mean score in its paper, we simply illustrate the mean score of AST.}.
For the multimodal input, we get the same score as PatchCore + FPFH method and is 1.6\% higher than the AST.
For single RGB input, we still have a 2\% improvement over PatchCore.
For 3D segmentation, similar to the AUPRO results reported in the main paper, our 3D segmentation results are a little bit lower than the FPFH-based method, and we believe this is also caused by the bias between the label and the point clouds we discuss in Section 4.7 in the main paper.
The P-AUROC is a saturated metric for anomaly segmentation, and the difference between methods is smaller than the difference in AUPRO.

\begin{table*}
  \centering
  \scriptsize
  \begin{tabular}{@{}cl|m{0.8cm}<{\centering}m{0.8cm}<{\centering}m{0.8cm}<{\centering}m{0.8cm}<{\centering}m{0.8cm}<{\centering}m{0.8cm}<{\centering}m{0.8cm}<{\centering}m{0.8cm}<{\centering}m{0.8cm}<{\centering}m{0.8cm}<{\centering}|m{0.8cm}<{\centering}}
    \toprule
    & Method & Bagel & Cable Gland & Carrot & Cookie & Dowel & Foam & Peach & Potato & Rope & Tire & Mean\\
    \midrule
     \multirow{2}*{\rotatebox{90}{3D}} &FPFH\cite{3d-ads} &    0.994&    0.966&    0.999&    0.946&    0.966& 0.927&    0.996&    0.999&    0.996&    0.990&    {\bf 0.978}\\
     ~&Ours &  0.981&  0.949&  0.997&  0.932&  0.959&  0.925&  0.989&  0.995&  0.994&  0.981&  0.970\\
    \midrule
     \multirow{2}*{\rotatebox{90}{RGB}} &PatchCore\cite{patchcore}&  0.983&  0.984&  0.980&  0.974&  0.972&  0.849&  0.976&  0.983&  0.987&  0.977&  0.967\\
     & Ours & 0.992 & 0.990 & 0.994 &  0.977 & 0.983 &  0.955 & 0.994 & 0.990 &  0.995 &   0.994& {\bf 0.987} \\
    \midrule
     \multirow{3}*{\rotatebox{90}{RGB+3D}} &AST\cite{ast} & -& -& -& -& -& -& -& -& -& -& 0.976  \\
     &PatchCore + FPFH\cite{3d-ads}& 0.996 & 0.992 & 0.997 & 0.994 & 0.981 & 0.974 & 0.996 & 0.998 & 0.994 & 0.995 & {\bf 0.992} \\
     &Ours & 0.995 &  0.993 & 0.997 & 0.985 & 0.985 & 0.984 & 0.996 & 0.994  & 0.997 & 0.996 & {\bf 0.992}\\
    \bottomrule
  \end{tabular}
  \caption{P-AUROC score for anomaly segmentation of all categories of MVTec-3D AD\cite{mvtec3dad} dataset. The P-AUROC is a saturated metric for anomaly segmentation, and the difference between methods is smaller than the AUPRO.}
  \label{tab:pauroc}
\end{table*}

\section{Detailed Results of Ablation Study}
\label{sec:ablation}

In the main paper Section 4.3, we conduct ablation studies on UFF, DLF, and multiple memory banks.
In this section, we report the detailed ablation study results of all categories of MVTec-3D AD.
\cref{tab:ablation_iauroc} and \cref{tab:ablation_aupro} separately  illustrate the I-AUPRO and AUPRO scores with the following settings:
1) Only Point Clouds ($\mathcal{M}_{pt}$) information; 
2) Only RGB ($\mathcal{M}_{rgb}$) information; 
3) Single memory bank ($\mathcal{M}_{fs}$) directly concatenating Point Transformer feature and RGB feature together;
4) Single memory bank ($\mathcal{M}_{fs}$) using UFF to fuse multimodal features;
5) Building two memory banks ($\mathcal{M}_{rgb},\mathcal{M}_{pt}$) separately and directly adding the scores together; 
6) Building two memory banks separately ($\mathcal{M}_{rgb},\mathcal{M}_{pt}$) and using DFL for the final result;
7) Building three memory banks ($\mathcal{M}_{rgb},\mathcal{M}_{pt},\mathcal{M}_{fs}$) (Ours).
With the UFF, the Foam, Cookie, and Peach have a great improvement to the single domain input and the w/o UFF version, which means the UFF encourages the interaction between multimodal features and creates useful information for anomaly detection and segmentation.
With double memory banks, the Carrot, Cookie and Potato score have an improvement, and the DLF help improve the hard categories such as Cable Gland and Tire.
With three memory banks, most advantages of DLF and UFF have been maintained, and our full setting gets the best results, which indicates DLF and UFF complements each other and jointly achieves the best performance. 

\begin{table*}
  \centering
  \scriptsize
  \begin{tabular}{@{}l|c|m{0.8cm}<{\centering}m{0.8cm}<{\centering}m{0.8cm}<{\centering}m{0.8cm}<{\centering}m{0.8cm}<{\centering}m{0.8cm}<{\centering}m{0.8cm}<{\centering}m{0.8cm}<{\centering}m{0.8cm}<{\centering}m{0.8cm}<{\centering}|m{0.8cm}<{\centering}}
    \toprule
     Method & Memory Banks & Bagel & Cable Gland & Carrot & Cookie & Dowel & Foam & Peach & Potato & Rope & Tire & Mean\\
    \midrule
     Only PC & $\mathcal{M}_{pt}$ & 0.941 & 0.651 & 0.965 & 0.969 & 0.905 & 0.760 & 0.880 & 0.974 & 0.926& 0.765 & 0.874\\
     Only RGB & $\mathcal{M}_{rgb}$ & 0.944 & 0.918 & 0.896 & 0.749 & 0.959 & 0.767 & 0.919 & 0.648 & 0.938 & 0.767 & 0.850 \\
     w/o  UFF & $\mathcal{M}_{fs}$ & 0.920 & 0.900 & 0.914  & 0.727 & 0.963 & 0.795 & 0.946 & 0.656 & 0.954 & 0.792 & 0.857 \\
     w/ UFF & $\mathcal{M}_{fs}$ & 0.976 & 0.895 & 0.922 & 0.912 & 0.949 & 0.868 & 0.978 & 0.723 & 0.960 & 0.798 & {\bf 0.898}\\
    \midrule
     w/o DLF & $\mathcal{M}_{pt}, \mathcal{M}_{rgb}$ & 0.981 & 0.831 & 0.980 & 0.985 & 0.960 & 0.905 & 0.936 & 0.964 & 0.967 & 0.780 & 0.929\\
     w/ DLF & $\mathcal{M}_{pt}, \mathcal{M}_{rgb}$ & 0.980 & 0.880 & 0.975 & 0.965 & 0.947 &0.910  & 0.943  & 0.927 & 0.958 & 0.840 & {\bf 0.932}\\
    \midrule
     Ours & $\mathcal{M}_{pt}, \mathcal{M}_{rgb}, \mathcal{M}_{fs}$ &  0.994 & 0.909 & 0.972 &  0.976 &  0.960 &  0.942 & 0.973 & 0.899& 0.972 & 0.850 & {\bf 0.945}\\
    \bottomrule
  \end{tabular}
  \caption{Detailed I-AUROC score for ablation on anomaly detection of all categories of MVTec-3D AD.}
  \label{tab:ablation_iauroc}
\end{table*}

\begin{table*}
  \centering
  \scriptsize
  \begin{tabular}{@{}l|c|m{0.8cm}<{\centering}m{0.8cm}<{\centering}m{0.8cm}<{\centering}m{0.8cm}<{\centering}m{0.8cm}<{\centering}m{0.8cm}<{\centering}m{0.8cm}<{\centering}m{0.8cm}<{\centering}m{0.8cm}<{\centering}m{0.8cm}<{\centering}|m{0.8cm}<{\centering}}
    \toprule
     Method & Memory Banks & Bagel & Cable Gland & Carrot & Cookie & Dowel & Foam & Peach & Potato & Rope & Tire & Mean\\
    \midrule
     Only PC & $\mathcal{M}_{pt}$ & 0.943 & 0.818 & 0.977 & 0.882 & 0.881 & 0.743 & 0.958 & 0.974 & 0.950 & 0.929 & 0.906\\
     Only RGB & $\mathcal{M}_{rgb}$ & 0.952 & 0.972& 0.973 & 0.891 & 0.932& 0.843 & 0.970 & 0.956& 0.968 & 0.966& 0.942\\
     w/o  UFF & $\mathcal{M}_{fs}$ & 0.951 & 0.971 & 0.974 & 0.893 & 0.935 & 0.855 & 0.972 & 0.958 & 0.969 & 0.967 & 0.944\\
     w/ UFF & $\mathcal{M}_{fs}$ &  0.963 & 0.964 & 0.978  &0.930  & 0.946 & 0.896 & 0.974 & 0.966 & 0.972 & 0.972 & {\bf 0.956}\\
    \midrule
     w/o DLF & $\mathcal{M}_{pt}, \mathcal{M}_{rgb}$ & 0.968 & 0.925 & 0.979 & 0.914 & 0.909 & 0.948 & 0.975 & 0.976 & 0.967 & 0.965 & 0.953\\
     w/ DLF & $\mathcal{M}_{pt}, \mathcal{M}_{rgb}$ & 0.965 & 0.968 & 0.978 & 0.933 & 0.933 & 0.927 & 0.976 & 0.967 & 0.971 & 0.973 & {\bf 0.959}\\
    \midrule
     Ours & $\mathcal{M}_{pt}, \mathcal{M}_{rgb}, \mathcal{M}_{fs}$ & 0.970 & 0.971 & 0.979 & 0.950 & 0.941 & 0.932 & 0.977 & 0.971  & 0.971 & 0.975 & {\bf 0.964}\\
    \bottomrule
  \end{tabular}
  \caption{Detailed AUPRO score for ablation anomaly segmentation of all categories of MVTec-3D AD.}
  \label{tab:ablation_aupro}
\end{table*}

\section{Detailed Results of PFA Analysis}
\label{sec:transformer}
In the main paper Section 4.4, we conduct experiments on PFA settings.
Here we report the detailed results of the main paper Table 5 with scores on each category in
\cref{tab:detail-pt-settiing} and \cref{tab:detail-pt-settiing-aupro}.
The PFA settings are:
\begin{itemize}
    \item Two important hyper-parameters of Point Transformer: the number of groups and the groups' size during farthest point sampling;
    The number of groups decides how many features will be extracted by the Point Transformer and the groups' size is equal to the concept of the receptive field;
    \item 3D anomaly detection experiment with the original point groups feature: a point group can be seen as a {\it patch}, and the memory bank store point groups feature here; 
    The detection method is as same as the patch-based one, and to get the segmentation predictions, we first project point group feature to a 2D plane and use an inverse distance interpolation to get every pixel value.
\end{itemize}

We found that directly calculating the anomaly on the point groups has some advantage in certain categories (e.g. Bagel, Cable Gland, Foam and Rope), and the reason is that after furthest point sampling (FPS) the original point group feature contains more small defects information.
As the patch gets smaller, our 2D plane point feature gets a better performance in detecting the small defects.

\begin{table*}
  \centering
  \scriptsize
  \begin{tabular}{@{}lcc|m{0.8cm}<{\centering}m{0.8cm}<{\centering}m{0.8cm}<{\centering}m{0.8cm}<{\centering}m{0.8cm}<{\centering}m{0.8cm}<{\centering}m{0.8cm}<{\centering}m{0.8cm}<{\centering}m{0.8cm}<{\centering}m{0.8cm}<{\centering}|m{0.8cm}<{\centering}}
    \toprule
     S.G & N.G & Sampling & Bagel & Cable Gland & Carrot & Cookie & Dowel & Foam & Peach & Potato & Rope & Tire & Mean\\
     \midrule
     64 & 784 & point groups & 0.933 & 0.579 & 0.854 & 0.843 & 0.874 & 0.748 & 0.761 & 0.863 & 0.989 & 0.483 & 0.793\\
     128 & 1024 & point groups & 0.945 & 0.690 & 0.905 & 0.925 & 0.897 & 0.809 & 0.854 & 0.888 & 0.991 & 0.509 & 0.841\\
     64 & 784 & $8\times8$ patch & 0.905 & 0.508 & 0.939 & 0.923 & 0.817 & 0.725 & 0.857 & 0.916 & 0.897 & 0.561 & 0.805 \\
     128 & 1024 & $8\times8$ patch & 0.886 & 0.560 & 0.925 & 0.971 & 0.832 & 0.711& 0.873 & 0.909 & 0.897 & 0.624 & 0.819\\
     128 & 1024 & $4\times4$ patch &  0.941 & 0.651 & 0.965 & 0.969 & 0.905 & 0.760 & 0.880 & 0.974 & 0.926& 0.765 & {\bf 0.874}\\
    \bottomrule
  \end{tabular}
  \caption{Detailed I-AUROC results of exploring Point Transformer setting on the pure 3D setting. S.G means the point number per group, and N.G means the total number of point groups.  We achieve the best performance with 1,024 point groups per sample and each point group contains 128 points; Compared with directly calculating anomaly scores on point groups, the method based on a 2D plane patch needs a small patch size towards high performance.}
  \label{tab:detail-pt-settiing}
\end{table*}

\begin{table*}
  \centering
  \scriptsize
  \begin{tabular}{@{}lcc|m{0.8cm}<{\centering}m{0.8cm}<{\centering}m{0.8cm}<{\centering}m{0.8cm}<{\centering}m{0.8cm}<{\centering}m{0.8cm}<{\centering}m{0.8cm}<{\centering}m{0.8cm}<{\centering}m{0.8cm}<{\centering}m{0.8cm}<{\centering}|m{0.8cm}<{\centering}}
    \toprule
     S.G & N.G & Sampling & Bagel & Cable Gland & Carrot & Cookie & Dowel & Foam & Peach & Potato & Rope & Tire & Mean\\
     \midrule
     64 & 784 & point groups & 0.906 & 0.709 & 0.942 & 0.854 & 0.869 & 0.681 & 0.871 & 0.906 & 0.943 & 0.447 & 0.813\\
     128 & 1024 & point groups & 0.956 & 0.812 & 0.964 & 0.895 & 0.892 & 0.644 & 0.965 & 0.974 & 0.966 & 0.891 & 0.896 \\
     64 & 784 & $8\times8$ patch & 0.899 & 0.789 &0.970 & 0.848 & 0.871 & 0.718 & 0.931 & 0.951 & 0.939 & 0.874 & 0.879 \\
     128 & 1024 & $8\times8$ patch & 0.934 & 0.808 & 0.977 & 0.856 & 0.877 & 0.745 & 0.949 & 0.970 & 0.948 & 0.894 & 0.896\\
     128 & 1024 & $4\times4$ patch & 0.943 & 0.818 & 0.977 & 0.882 & 0.881 & 0.743 & 0.958 & 0.974 & 0.950 & 0.929 & {\bf 0.906}\\
    \bottomrule
  \end{tabular}
  \caption{Detailed AUPRO results of exploring Point Transformer setting on the pure 3D setting. S.G means the point number per group, and N.G means the total number of point groups.  We get the best performance with 1024 point groups per sample and each point group contains 128 points; Compared with directly calculating segmentation scores on point groups, the method based on a 2D plane patch needs a small patch size towards high performance.}
  \label{tab:detail-pt-settiing-aupro}
\end{table*}

\section{Detailed Results of Few-shot Setting}
\label{sec:few-shot}

In the main paper Section 4.6, we evaluate our method on Few-shot settings, and the detailed results on all of the categories are illustrated in \cref{tab:detail-fs-au} and \cref{tab:detail-fs-iauroc}.
We randomly select 10 and 5 images from each category as training data and test the few-shot model on the full testing dataset.
We find that our method in a 10-shot or 5-shot setting still has a better segmentation performance than some non-few-shot methods.
In the 50-shot setting, We found that some categories get better performance than the full dataset version (e.g. Bagel and Potato), which means the memory bank building algorithm still has some improvement space, and we will discuss the problem in future research.

\begin{table*}
  \centering
  \scriptsize
  \begin{tabular}{@{}l|m{0.8cm}<{\centering}m{0.8cm}<{\centering}m{0.8cm}<{\centering}m{0.8cm}<{\centering}m{0.8cm}<{\centering}m{0.8cm}<{\centering}m{0.8cm}<{\centering}m{0.8cm}<{\centering}m{0.8cm}<{\centering}m{0.8cm}<{\centering}|m{0.8cm}<{\centering}}
    \toprule
     Method & Bagel & Cable Gland & Carrot & Cookie & Dowel & Foam & Peach & Potato & Rope & Tire & Mean\\
     \midrule
     5-shot  &0.974 &	0.645&	0.833&	0.942&	0.636&	0.798&	0.820&	0.781&	0.914&	0.615&	0.796\\
     10-shot &0.987&	0.662&	0.854&	0.969&	0.643&	0.799&	0.908&	0.771&	0.931&	0.682&	0.821\\
     50-shot &0.997&	0.745&	0.957&	0.966&	0.910&	0.915&	0.937&	0.910&	0.946&	0.744&	0.903\\
     Full dataset  &  0.994 & 0.909 & 0.972 &  0.976 &  0.960 &  0.942 & 0.973 & 0.899& 0.972 & 0.850 &  {\bf 0.945}\\
    \bottomrule
  \end{tabular}
  \caption{Few-shot I-AUROC of all categories of MVTec-3D AD. Our method still has good anomaly detection performance on few-shot settings.}
  \label{tab:detail-fs-iauroc}
\end{table*}

\begin{table*}
  \centering
  \scriptsize
  \begin{tabular}{@{}l|m{0.8cm}<{\centering}m{0.8cm}<{\centering}m{0.8cm}<{\centering}m{0.8cm}<{\centering}m{0.8cm}<{\centering}m{0.8cm}<{\centering}m{0.8cm}<{\centering}m{0.8cm}<{\centering}m{0.8cm}<{\centering}m{0.8cm}<{\centering}|m{0.8cm}<{\centering}}
    \toprule
     Method & Bagel & Cable Gland & Carrot & Cookie & Dowel & Foam & Peach & Potato & Rope & Tire & Mean\\
     \midrule
     5-shot &0.959&	0.879&	0.974&	0.906&	0.879&	0.848&	0.968&	0.957&	0.963&	0.935&	0.927\\
     10-shot &0.972& 0.910&	0.976&	0.923&	0.905&	0.870&	0.972&	0.956&	0.967&	0.939&	0.939\\
     50-shot &0.969&	0.955&	0.977&	0.940&	0.906&	0.912&	0.971&	0.965&	0.968&	0.959&	0.952\\
     Full dataset & 0.970 & 0.971 & 0.979 & 0.950 & 0.941 & 0.932 & 0.977 & 0.971  & 0.971 & 0.975 & {\bf 0.964}\\
    \bottomrule
  \end{tabular}
  \caption{Few-shot AUPRO of all categories of MVTec-3D AD. Our method on few-shot still has a better anomaly segmentation performance than most non-few-shot methods.}
  \label{tab:detail-fs-au}
\end{table*}

\section{Backbone Choices}
\label{sec:backbone}

Feature extractors play an important role in anomaly detection. In this section, we explore different backbone settings on both point cloud and RGB images. For RGB we compare four extractor settings:
1) A ViT-B/8 supervised backbone pretrained with ImageNet\cite{imagenet} 1K;
2) A ViT-B/8 supervised backbone pretrained with ImageNet 21K;
3) A ViT-S/8 self-supervised backbone pertrained via DINO\cite{DINO};
4) A ViT-B/8 self-supervised backbone pertrained via DINO.
And for Point Clouds transformer, we compare two self-supervised pretrained backbones:
1) Point-Bert\cite{point-bert}; 2) Point-MAE\cite{pointmae}.
The detection and segmentation results are separately illustrated in \cref{tab:backbone_iauroc} and \cref{tab:backbone_aupro}.
The results show that  
the self-supervised pretrained methods have better results than the supervised ones, and 
small backbones pretrained with self-supervised methods perform better than the bigger ones pretrained with supervised methods.
The performance of Point-MAE is better than that of Point-Bert, we think the reason is that Point-MAE needs to reconstruct more point cloud details than Point-Bert, thus can catch small defects in anomaly detection.

\begin{table*}
  \centering
  \scriptsize
  \begin{tabular}{@{}cl|m{0.8cm}<{\centering}m{0.8cm}<{\centering}m{0.8cm}<{\centering}m{0.8cm}<{\centering}m{0.8cm}<{\centering}m{0.8cm}<{\centering}m{0.8cm}<{\centering}m{0.8cm}<{\centering}m{0.8cm}<{\centering}m{0.8cm}<{\centering}|m{0.8cm}<{\centering}}
    \toprule
    & Method & Bagel & Cable Gland & Carrot & Cookie & Dowel & Foam & Peach & Potato & Rope & Tire & Mean\\
    \midrule
     \multirow{4}*{\rotatebox{90}{RGB}} & Supervised ImageNet 1K&0.793	&0.729&	0.774&	0.709&	0.723&	0.601&	0.607&	0.606&	0.605&	0.556&	0.670\\
     &Supervised ImageNet 21K  & 0.814& 	0.658&	0.788&	0.630&	0.784&	0.582&	0.615&	0.459&	0.674&	0.621&	0.662\\
     &DINO ViT-S/8  & 0.933&	0.865&	0.898&	0.786&	0.878&	0.759&	0.902&	0.520 &	0.898&	0.748&	0.819\\
     & DINO ViT-B/8 & 0.944 & 0.918 & 0.896 & 0.749 & 0.959 & 0.767 & 0.919 & 0.648 & 0.938 & 0.767 & {\bf 0.850} \\
     \midrule
     \multirow{2}*{\rotatebox{90}{3D}} & Point-Bert & 0.900&	0.632&	0.932&	0.915&	0.851&	0.659&	0.826&	0.899&	0.894&	0.530&	0.803\\
     &Point-MAE  & 0.941 & 0.651 & 0.965 & 0.969 & 0.905 & 0.760 & 0.880 & 0.974 & 0.926& 0.765 & {\bf 0.874}\\
    \bottomrule
  \end{tabular}
  \caption{I-AUROC score for anomaly detection of MVTec-3D AD\cite{mvtec3dad} dataset with different backbone. For RGB feature extractor, The self-supervised backbone is better than the supervised ones.}
  \label{tab:backbone_iauroc}
\end{table*}

\begin{table*}
  \centering
  \scriptsize
  \begin{tabular}{@{}cl|m{0.8cm}<{\centering}m{0.8cm}<{\centering}m{0.8cm}<{\centering}m{0.8cm}<{\centering}m{0.8cm}<{\centering}m{0.8cm}<{\centering}m{0.8cm}<{\centering}m{0.8cm}<{\centering}m{0.8cm}<{\centering}m{0.8cm}<{\centering}|m{0.8cm}<{\centering}}
    \toprule
    & Method & Bagel & Cable Gland & Carrot & Cookie & Dowel & Foam & Peach & Potato & Rope & Tire & Mean\\
    \midrule
     \multirow{4}*{\rotatebox{90}{RGB}} & Supervised ImageNet 1K& 0.844&	0.842&	0.892&	0.681&	0.842&	0.568&	0.765&	0.865&	0.915&	0.871&	0.808\\
     & Supervised ImageNet 21K &0.805&	0.878&	0.927&	0.712&	0.888&	0.62&	0.785&	0.909&	0.919&	0.930&	0.837\\
     & DINO ViT-S/8 & 0.948	&0.973&	0.971&	0.906&	0.947&	0.788&	0.972&	0.954&	0.964&	0.949&	0.937\\
     & DINO ViT-B/8 &0.952&	0.972&	0.973&	0.891&	0.932&	0.843&	0.970&	0.956&	0.968&	0.966&	{\bf 0.942}\\
     \midrule
     \multirow{2}*{\rotatebox{90}{3D}} & Point-Bert & 0.895 &	0.775&	0.972&	0.841&	0.871&	0.680&	0.918&	0.964&	0.938&	0.877&	0.873\\
     & Point-MAE & 0.943&	0.818&	0.977&	0.882&	0.881&	0.743&	0.958&	0.974&	0.950&	0.929&	{\bf 0.906}\\
    \bottomrule
  \end{tabular}
  \caption{AUPRO score for anomaly segmentation of MVTec-3D AD\cite{mvtec3dad} dataset with different backbones. For RGB feature extractor, the self-supervised backbone is better than the supervised ones.}
  \label{tab:backbone_aupro}
\end{table*}

\section{Eyecandies Results}
\label{sec:eyecandies}

We have noticed that recently a new dataset Eyecandies~\cite{eyecandies} provides multimodel information of 10 categories of candies, and each category contains 1000 images for training and 50 images for public testing. 
The source dataset provides 6 RGB images, which are in different light conditions, a depth map, and a normal map of each sample.
In this section, we convert the Eyecandies dataset to the format supported by  M3DM.
In detail, we use the environment light image as our input RGB data, and for 3D data, we first convert the depth image to point clouds with internal parameters, then we remove the background points with point coordinates.
For computation efficiency, we use only less than 400 samples from each category for training.
We try to build memory banks of different sizes (ranging from 10 to 400 samples) to find the best one under this dataset.
As illustrated in \cref{tab:eyecandies} and \cref{tab:eyecandies_p_aucroc}, we report the best I-AUCROC and P-AUCROC scores.
Compared with baseline methods, we have significant improvement in both the RGB setting and RGB+3D setting.
Previous work did not report the AUPRO score on the Eyecandies dataset, and for reference in further study, we provide the this segmentation performance metric score of M3DM in  \cref{tab:eyecandies_aupro}.

\begin{table*}
  \centering
  \scriptsize
  \begin{tabular}{@{}cl|m{0.8cm}<{\centering}m{0.8cm}<{\centering}m{0.8cm}<{\centering}m{0.8cm}<{\centering}m{0.8cm}<{\centering}m{0.8cm}<{\centering}m{0.8cm}<{\centering}m{0.8cm}<{\centering}m{0.8cm}<{\centering}m{0.8cm}<{\centering}|m{0.8cm}<{\centering}}
    \toprule
    & Method & Candy Cane & Chocolate Cookie & Chocolate Praline & Confetto & Gummy Bear & Hazelnut Truffle & Licorice Sandwich & Lollipop & Marshm -allow & Pepper-mint Candy & Mean\\
    \midrule
     \multirow{1}*{\rotatebox{90}{3D}} &Ours  &   0.482  & 0.589 & 0.805 & 0.845 & 0.780 & 0.538 & 0.766 & 0.827 & 0.800 & 0.822 & 0.725\\
    \midrule
     \multirow{4}*{\rotatebox{90}{RGB}} & RGB\cite{eyecandies}& 0.527 & 0.848 & 0.772 & 0.734 & 0.590 & 0.508 & 0.693 & 0.760 & 0.851 & 0.730 & 0.701 \\
     & STFPM\cite{STFPM} &  0.551 & 0.654 & 0.576 & 0.784 & 0.737 & \bf 0.790 & 0.778 & 0.620 & 0.840 & 0.749 & 0.708\\
     & PaDiM\cite{padim} & 0.531 & 0.816 & 0.821 & 0.856 & 0.826 & 0.727 & 0.784 & 0.665 & 0.987 & 0.924 & 0.794 \\
     & Ours & \bf 0.648  & \bf	0.949  & \bf	0.941 & \bf	1.000	 & \bf 0.878 & 	0.632 & \bf	0.933 & \bf	0.811 & \bf	0.998 & \bf	1.000 & \bf	0.879 \\
    \midrule
     \multirow{3}*{\rotatebox{90}{RGB+3D}} & RGB-D\cite{eyecandies} & 0.529 & 0.861 & 0.739 & 0.752 & 0.594 & 0.498 & 0.679 & 0.651 & 0.838 & 0.75 & 0.689 \\
     &RGB-cD-N\cite{eyecandies} & 0.596 & 0.843 & 0.819 & 0.846 & 0.833 & 0.550 & 0.750 & \bf 0.846 & 0.940 & 0.848 & 0.787  \\
     &Ours  & \bf 0.624 & \bf 0.958 & \bf 0.958 & \bf 1.000 & \bf 0.886 & \bf 0.758 & \bf 0.949 &  0.836 &  \bf 1.000 & \bf  1.000 & \bf 0.897\\ 
    \bottomrule
  \end{tabular}
  \caption{I-AUROC score for anomaly detection of all categories of Eyecandies~\cite{eyecandies} dataset. The results of baselines are from the ~\cite{eyecandies}.}
  \label{tab:eyecandies}
\end{table*}

\begin{table*}
  \centering
  \scriptsize
  \begin{tabular}{@{}cl|m{0.8cm}<{\centering}m{0.8cm}<{\centering}m{0.8cm}<{\centering}m{0.8cm}<{\centering}m{0.8cm}<{\centering}m{0.8cm}<{\centering}m{0.8cm}<{\centering}m{0.8cm}<{\centering}m{0.8cm}<{\centering}m{0.8cm}<{\centering}|m{0.8cm}<{\centering}}
    \toprule
    & Method & Candy Cane & Chocolate Cookie & Chocolate Praline & Confetto & Gummy Bear & Hazelnut Truffle & Licorice Sandwich & Lollipop & Marshm -allow & Pepper-mint Candy & Mean\\
    \midrule
     \multirow{1}*{\rotatebox{90}{3D}} & Ours & 0.977 & 0.903 & 0.902 & 0.93 & 0.875 & 0.832 & 0.909 & 0.968 & 0.868 & 0.918 & 0.908 \\
    \midrule
     \multirow{2}*{\rotatebox{90}{RGB}} & RGB\cite{eyecandies}& \bf 0.972 & 0.933 & \textbf{0.960} & 0.945 & 0.929 & 0.815 & 0.855 & 0.977 & 0.931 & 0.928 & 0.925 \\
     & Ours & \textbf{0.956} & \textbf{0.979} & 0.958 & \textbf{0.998} & \textbf{0.976} & \textbf{0.941} & \textbf{0.977} & \textbf{0.986} & \textbf{0.997} & \textbf{0.988} & \textbf{0.976} \\
    \midrule
     \multirow{3}*{\rotatebox{90}{RGB+3D}} & RGB-D\cite{eyecandies} & 0.973          & 0.927          & 0.958          & 0.945          & 0.929          & 0.806          & 0.827          & 0.977          & 0.931          & 0.928          & 0.920          \\
     &RGB-cD-N\cite{eyecandies} & \bf 0.980          & 0.979          & \bf 0.982          & 0.978          & 0.951          & 0.853          & 0.971          & 0.978          & 0.985          & 0.967          & 0.962   \\
     &Ours  & 0.974          & \textbf{0.987} & 0.962          & \textbf{0.998} & \textbf{0.966} & \textbf{0.941} & \textbf{0.973} & \textbf{0.984} & \textbf{0.996} & \textbf{0.985} & \textbf{0.977} \\
    \bottomrule
  \end{tabular}
  \caption{P-AUROC score for anomaly detection of all categories of Eyecandies~\cite{eyecandies} dataset. The results of baselines are from the ~\cite{eyecandies}.}
  \label{tab:eyecandies_p_aucroc}
\end{table*}

\begin{table*}
  \centering
  \scriptsize
  \begin{tabular}{l|m{0.8cm}<{\centering}m{0.8cm}<{\centering}m{0.8cm}<{\centering}m{0.8cm}<{\centering}m{0.8cm}<{\centering}m{0.8cm}<{\centering}m{0.8cm}<{\centering}m{0.8cm}<{\centering}m{0.8cm}<{\centering}m{0.8cm}<{\centering}|m{0.8cm}<{\centering}}
    \toprule
     Method & Candy Cane & Chocolate Cookie & Chocolate Praline & Confetto & Gummy Bear & Hazelnut Truffle & Licorice Sandwich & Lollipop & Marshm -allow & Pepper-mint Candy & Mean\\
     \midrule
     Point Clouds &0.911          & 0.645          & 0.581          & 0.748          & 0.748          & 0.484          & 0.608          & 0.904          & 0.646          & 0.750           & 0.702  \\
     RGB & 0.867          & 0.904          & 0.805          & 0.982          & 0.871          & 0.662          & 0.882          & 0.895          & 0.970           & 0.962          & 0.880 \\
     Point Clouds + RGB & 0.906          & 0.923          & 0.803          & 0.983          & 0.855          & 0.688          & 0.880          & 0.906          & 0.966          & 0.955          & 0.882 \\
    \bottomrule
  \end{tabular}
  \caption{AUPRO score for anomaly detection of all categories of Eyecandies~\cite{eyecandies} dataset.}
  \label{tab:eyecandies_aupro}
\end{table*}

\section{Visualization Results}
\label{sec:visual}

\begin{figure*}[t]
  \centering
  \includegraphics[width=0.9\linewidth]{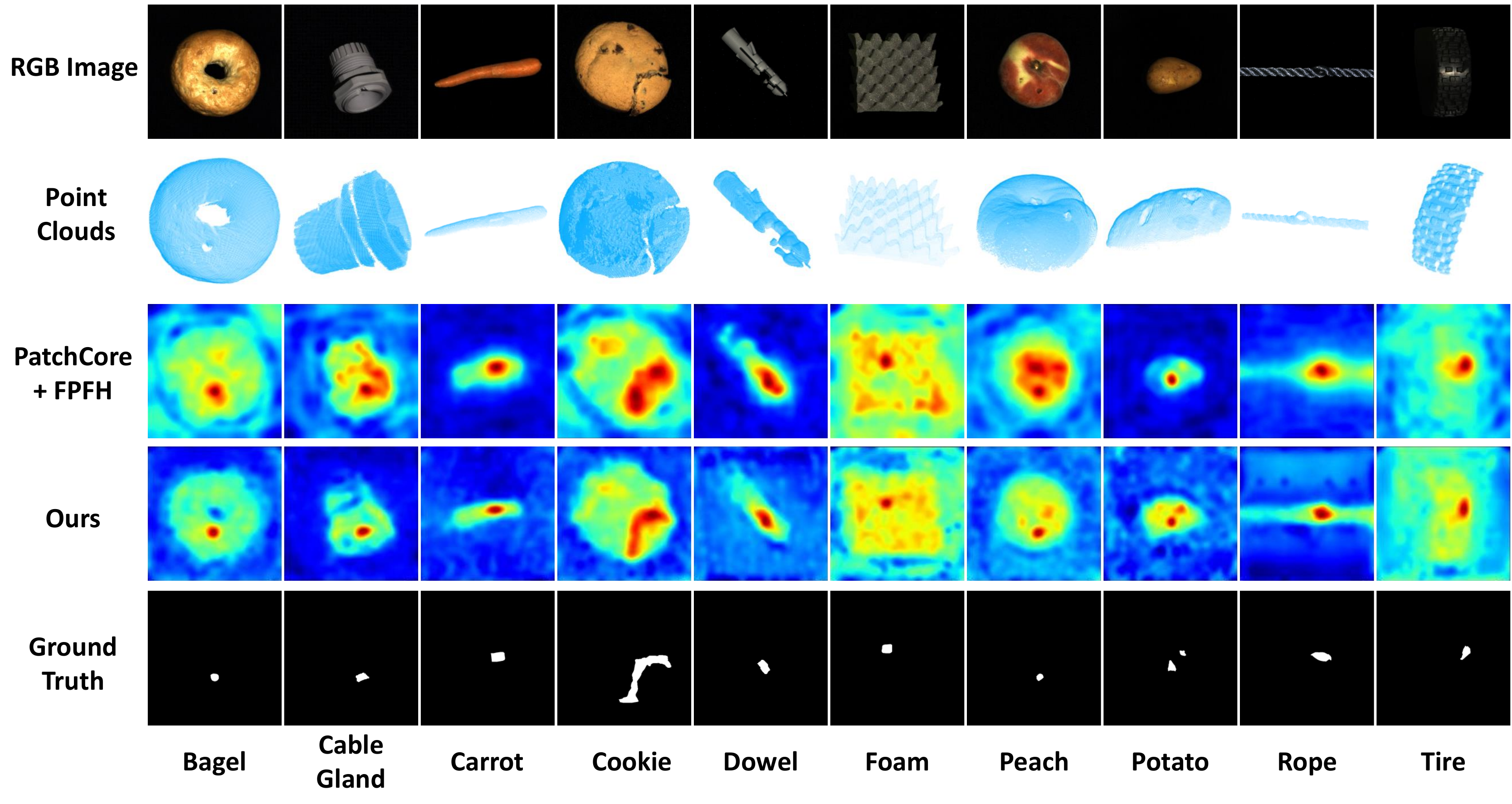}

   \caption{Heatmap of our anomaly segmentation results (multimodal inputs). Compared with PatchCore + FPFH, our method outputs a more accurate segmentation region.}
   \label{fig:compare}
\end{figure*}

In this section, we visualize more anomaly segmentation results for all categories of MVTec-3D AD datasets.
As shown in \cref{fig:compare}, we visualize the heatmap results of our method and PatchCore + FPFH, both with multimodal inputs. Compared with PatchCore + FPFH results, our method gets better segmentation maps. 

\end{subappendices}
\end{document}